\documentclass[10pt,twocolumn,letterpaper]{article}

\usepackage{cvpr}              

\usepackage{graphicx}
\usepackage{amsmath}
\usepackage{amssymb}
\usepackage{booktabs}

\usepackage{floatrow}
\usepackage{tabularx}
\usepackage{makecell}
\usepackage{booktabs}
\usepackage{multirow}
\usepackage{amsmath}
\usepackage{xcolor}
\usepackage{color, colortbl}
\usepackage[export]{adjustbox}
\definecolor{Gray}{gray}{0.92}

\usepackage[pagebackref,breaklinks,colorlinks]{hyperref}

\usepackage[capitalize]{cleveref}
\crefname{section}{Sec.}{Secs.}
\Crefname{section}{Section}{Sections}
\Crefname{table}{Table}{Tables}
\crefname{table}{Tab.}{Tabs.}

\def\cvprPaperID{*****} 
\def\confName{CVPR}
\def\confYear{2022}

\newcommand\blfootnote[1]{%
  \begingroup
  \renewcommand\thefootnote{}\footnote{#1}%
  \addtocounter{footnote}{-1}%
  \endgroup
}

\newcommand{\figshortref}[1]{Fig.~\ref{#1}}
\newcommand{\tabref}[1]{Table~\ref{#1}}
\newcommand{\tabfref}[1]{Table~\ref{#1}}

\newcommand{\eqnref}[1]{(\ref{#1})}
\newcommand{\secref}[1]{Section~\ref{#1}}

\newcommand\scalemath[2]{\scalebox{#1}{\mbox{\ensuremath{\displaystyle #2}}}}

\DeclareMathOperator*{\argmax}{argmax}

\begin{document}

\title{Source domain subset sampling for semi-supervised domain adaptation in semantic segmentation}

\author{Daehan Kim{$^{\dagger}$}\\
Hanbat National University\\
{\tt\small daehan.kim@edu.hanbat.ac.kr}
\and
Minseok Seo{$^{\dagger}$}\\
Hanbat National University\\
{\tt\small minseok.seo@edu.hanbat.ac.kr}
\and
Jinsun Park{$^{\dagger}$}\\
Pusan National University\\
{\tt\small jspark@pusan.ac.kr}
\and
Dong-Geol Choi{$^{*}$}\\
Hanbat National University\\
{\tt\small dgchoi@hanbat.ac.kr}
}
\maketitle

\blfootnote{$\dagger$ \text{These authors contributed equally.}}

\begin{abstract}
In this paper, we introduce source domain subset sampling (SDSS) as a new perspective of semi-supervised domain adaptation. 
We propose domain adaptation by sampling and exploiting only a meaningful subset from source data for training. 
Our key assumption is that the entire source domain data may contain samples that are unhelpful for the adaptation.
Therefore, the domain adaptation can benefit from a subset of source data composed solely of helpful and relevant samples.
The proposed method effectively subsamples full source data to generate a small-scale meaningful subset. Therefore, training time is reduced, and performance is improved with our subsampled source data.
To further verify the scalability of our method, we construct a new dataset called \textit{Ocean Ship}, which comprises 500 real and 200K synthetic sample images with ground-truth labels.
The SDSS achieved a state-of-the-art performance when applied on GTA5 → Cityscapes and SYNTHIA → Cityscapes public benchmark datasets and a 9.13 mIoU improvement on our \textit{Ocean Ship} dataset over a baseline model. 
\end{abstract}

\section{Introduction}
\label{sec:intro}
Semantic segmentation is one of the most important tasks in computer vision, which estimates the pixel-level semantic labels of an image.
The performance of segmentation has been substantially improved with the recent advances in deep learning~\cite{long2015fully, ronneberger2015u, zhao2017pyramid, chen2017deeplab, chen2018encoder, SunZJCXLMWLW19, huang2019ccnet}.
However, existing methods need to be trained on large-scale segmentation datasets~\cite{cordts2016cityscapes, lin2014microsoft, alhaija2018augmented}, where the acquisition of accurate ground-truth (GT) labels often requires massive human labour and high costs.
As a proxy to a real-world dataset, synthetic datasets, such as GTA5~\cite{richter2016playing} and SYNTHIA~\cite{ros2016synthia}, have been proposed. 
These synthetic datasets provide images generated by various simulations, together with highly accurate pixel-level GT segmentation annotations.
Although they have successfully resolved the data deficiency problem, a critical issue remains as the network trained on the source (i.e., synthetic) dataset poorly generalizes on the target (i.e., real-world) dataset due to the large domain gap between the datasets.

\begin{figure*}[!t]
    \centering
    \def\arraystretch{0.1}
    \begin{tabular}{@{\hskip 0.003\linewidth}c@{\hskip 0.003\linewidth}c@{\hskip 0.003\linewidth}c}
    \includegraphics[width=0.33\linewidth]{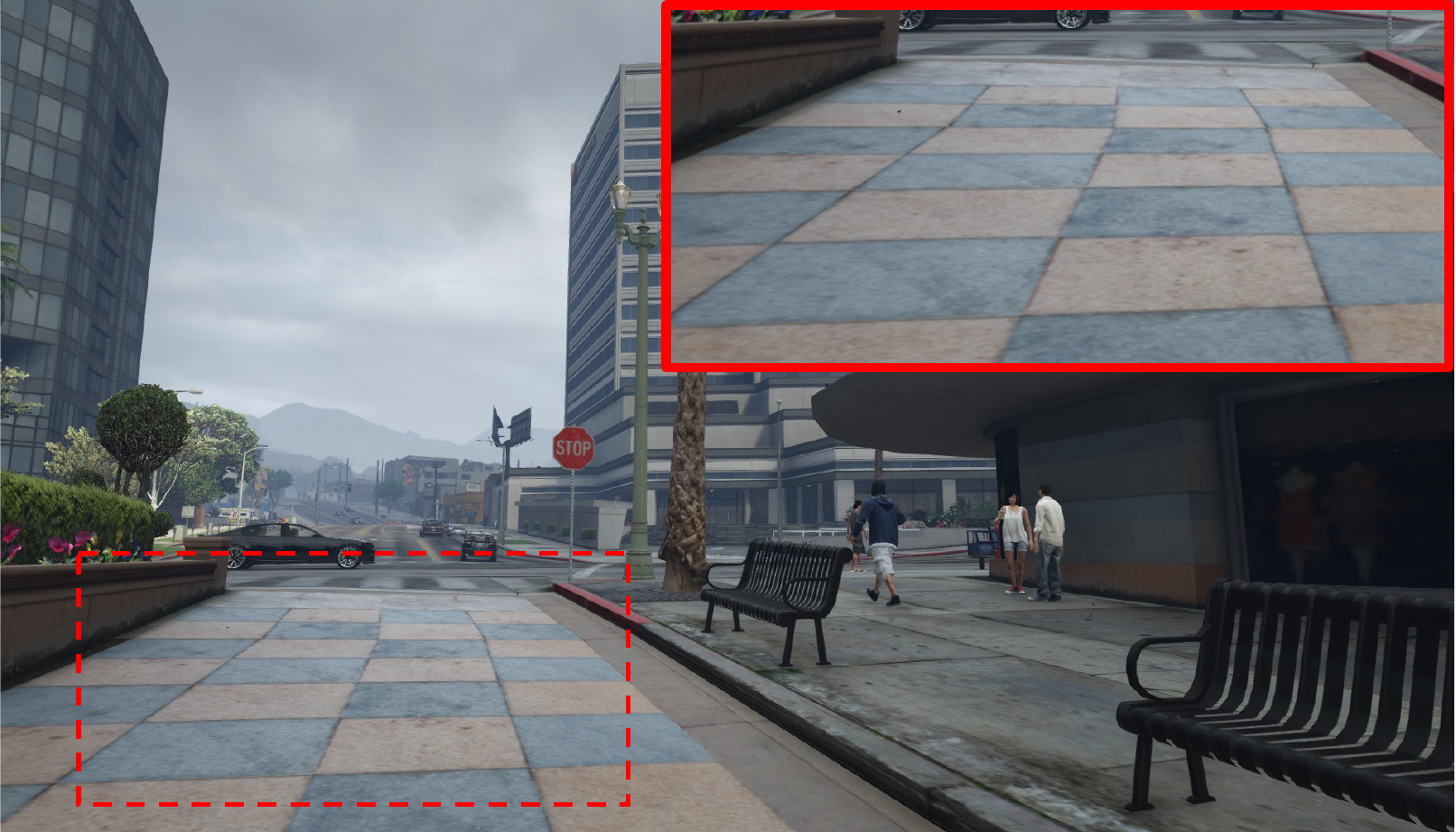}&
    \includegraphics[width=0.33\linewidth]{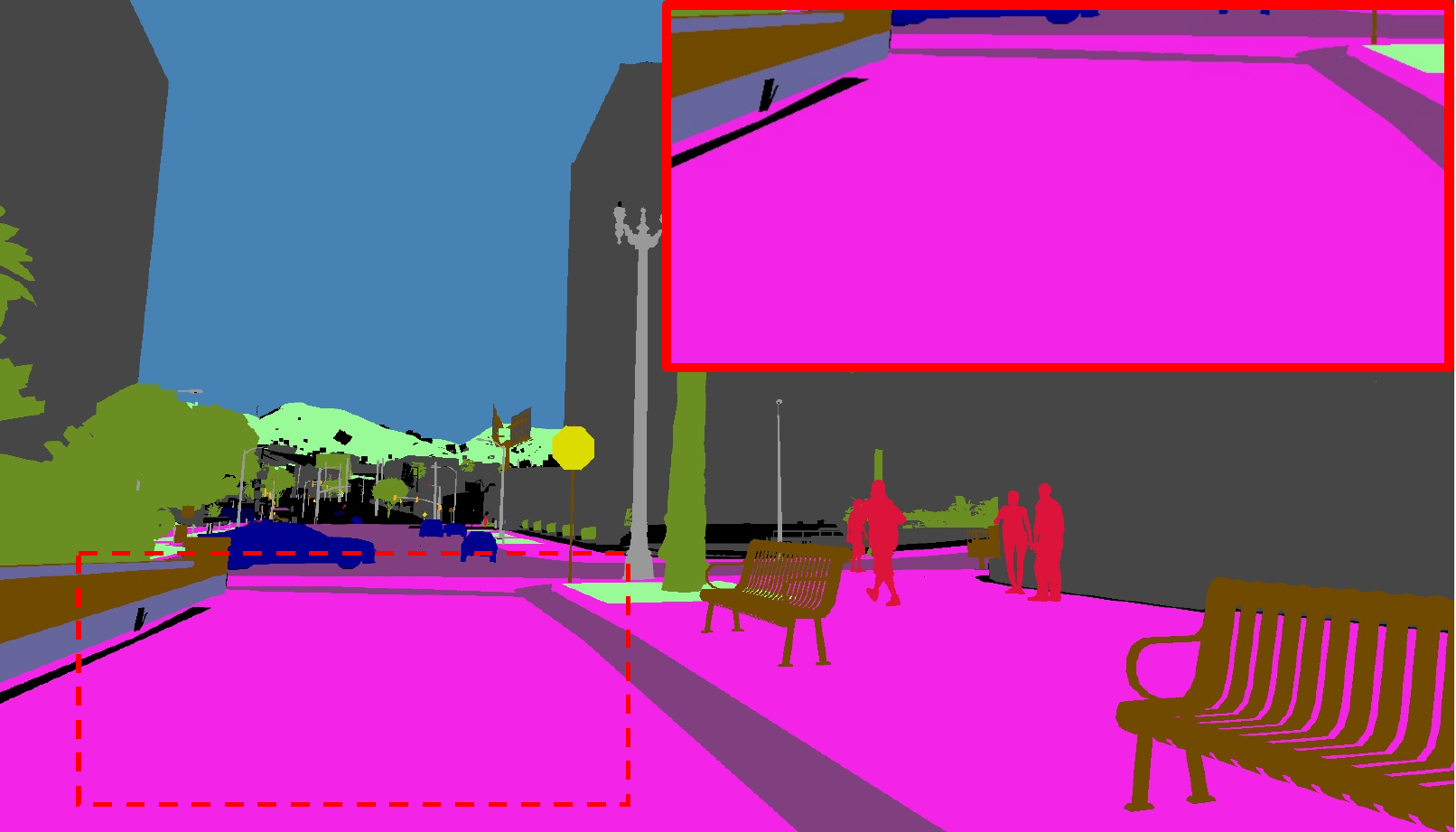} &
    \includegraphics[width=0.33\linewidth]{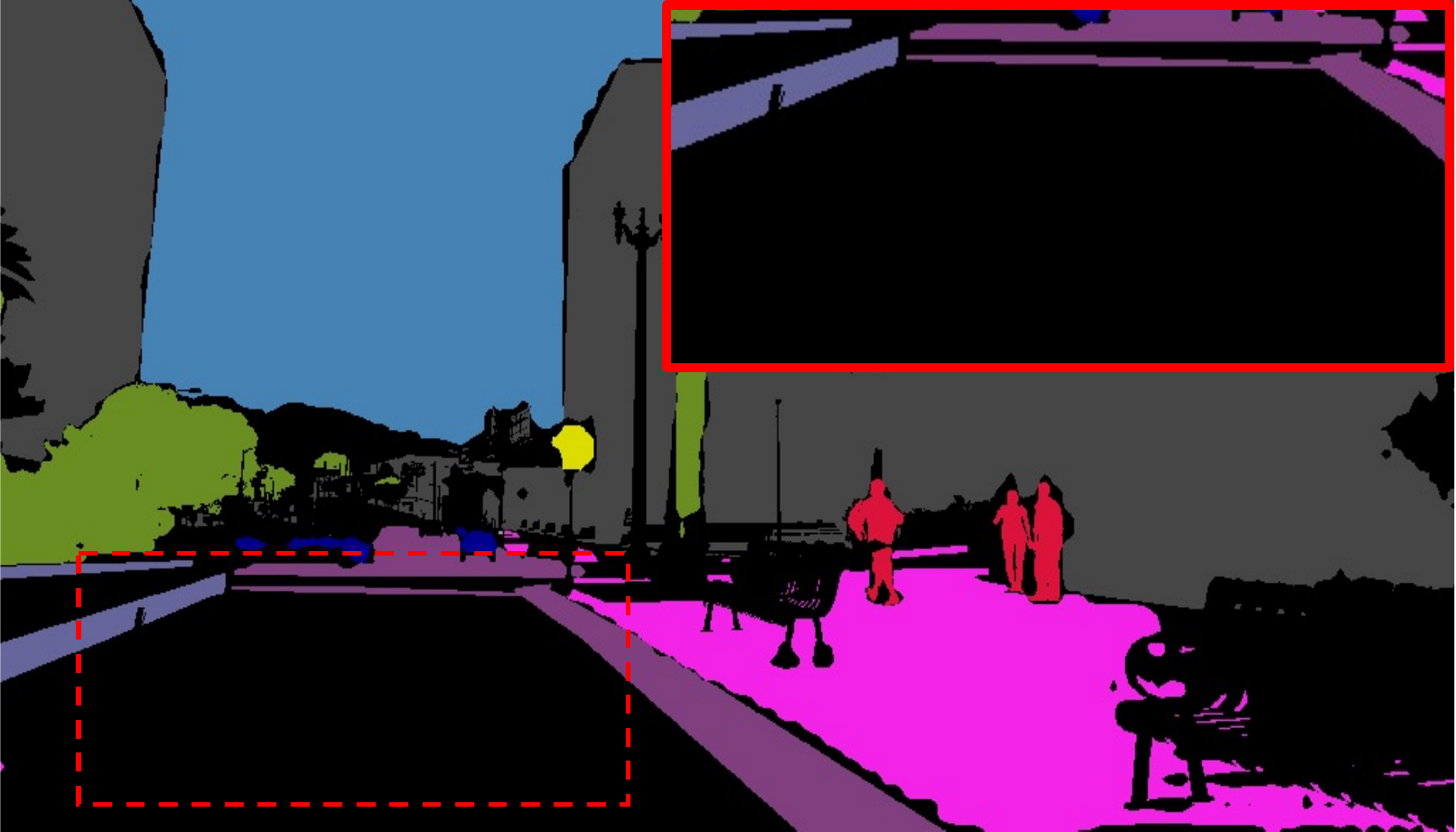} \\
    \vspace{+0.05mm}&\vspace{+0.05mm}&\vspace{+0.05mm}\\
    \includegraphics[width=0.33\linewidth]{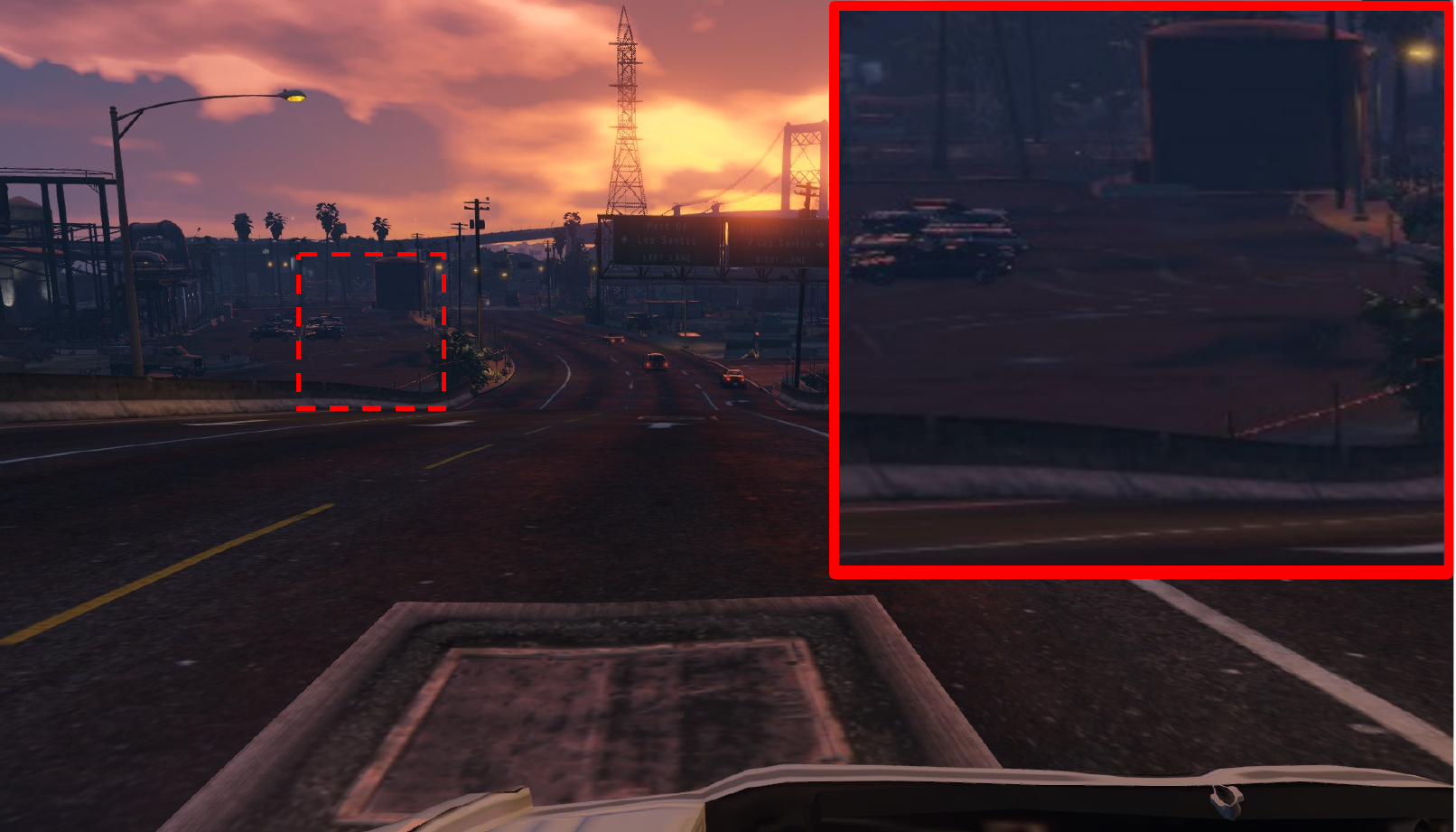}&
    \includegraphics[width=0.33\linewidth]{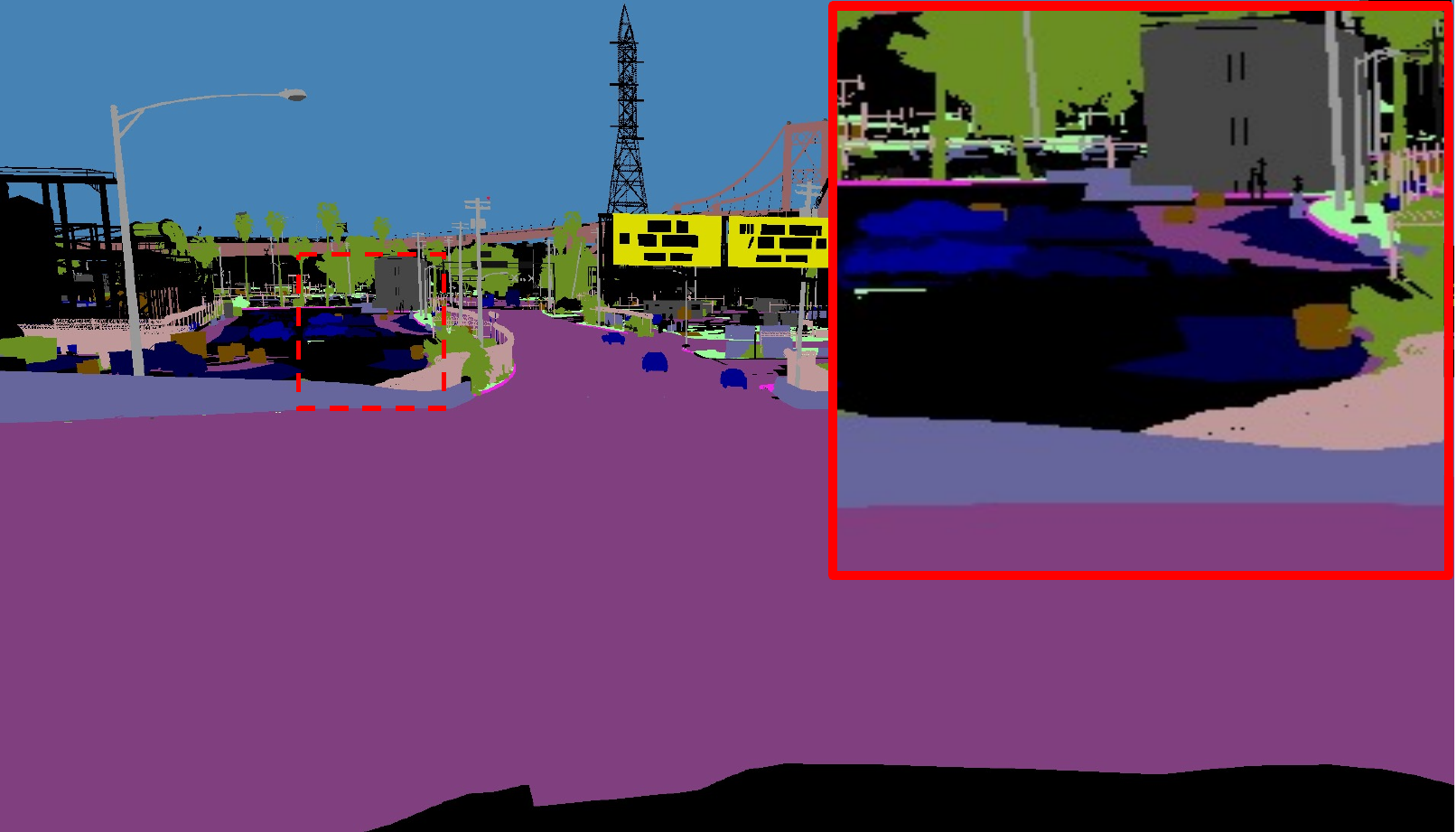}&
    \includegraphics[width=0.33\linewidth]{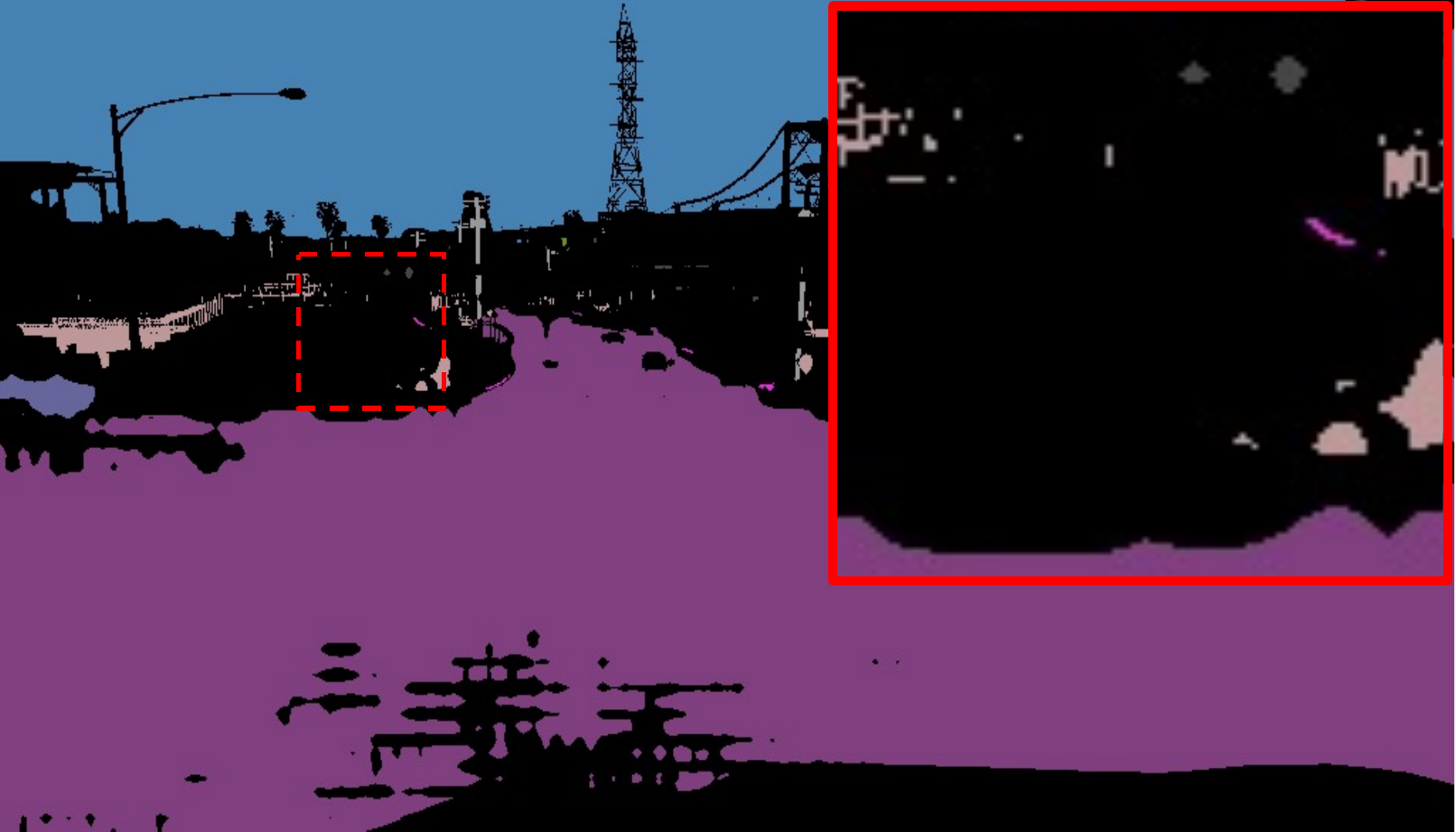} \\
    \vspace{+0.15mm}&\vspace{+0.15mm}&\vspace{+0.15mm}\\
    \scriptsize{(a) Source image} & \scriptsize{(b) GT} & \scriptsize{(c) Ours}
    \end{tabular}
    \caption{
    Sampling results on GTA5 (source) with the proposed SDSS method. Our method successfully eliminates confusing GT labels based on the knowledge of the Cityscapes (target) dataset.}
    \label{figure:figure1} 
\end{figure*}

Domain adaptation (DA) has drawn the attention of researchers to reduce the domain gap between the source and target images.
In particular, unsupervised domain adaptation (UDA) methods~\cite{zou2018unsupervised, zhang2019category, mei2020instance} aim to resolve the domain gap problem with the source and target images without any GT of the target images. Although UDA methods have shown promising results, it is still challenging to overcome the problem and achieve comparable performance to that of fully supervised methods~\cite{minaee2021image} without any reliable guidance from the target domain.
Moreover, semi-supervised domain adaptation (SSDA) methods exploit not only source domain data but also a few labelled target domain data during training~\cite{saito2019semi, jiang2020bidirectional, kim2020attract, wang2020alleviating}.
In this paper, we propose a new SSDA approach to eliminate irrelevant source domain samples and benefit from a subset of source domain data only containing samples highly relevant to the target domain.
We argue that there exist irrelevant or even unhelpful samples in large-scale source domain data generated even with a carefully controlled data generation scenario.
\figshortref{figure:figure1} shows examples of confusing source domain samples.
Because of the unconstrained movements during the simulation, a vehicle is seen driving on the sidewalk (c.f, 1st row).
Humans often perceive this area as a road because of its similar appearance to a road and their basic knowledge of its geometry.
These types of confusing source domain samples may hinder the effective domain adaptation~\cite{choi2020cars}. 
In other words, this phenomenon often results in confusing training samples that may mislead the training procedure.

Therefore, we present the source domain subset sampling (SDSS) method to exploit a subset of samples selected from the entire source domain data during training. 
The subset sampled through the proposed method is more relevant to the target domain, showing synergistic effects when combined with previous methods~\cite{saito2019semi, wang2020alleviating}.
The proposed approach eliminates the aforementioned confusing samples and ensures that the training process facilitates only relevant and useful samples, as depicted in \figshortref{figure:figure1} (c).

We first utilized a small number of target data with GT labels to pre-train a segmentation network.
Then, we extracted pixel-level source samples from a source domain by eliminating source samples incorrectly predicted by the pre-trained network.
Furthermore, we selected image-level source samples by simultaneously considering class balance and prediction correctness within an image.
With our approach, a subset of source data providing class-balanced and target domain relevant information was obtained efficiently.
Experimental results on the GTA5 and SYNTHIA datasets show that our method is effective in achieving superior segmentation performance with a reduced number of source data and shorter training time.
We further presented a new \textit{Ocean Ship} dataset containing 500 real and 200K synthetic images and validated the proposed method on our dataset. Our contributions can be summarized as follows:

\begin{figure*}[t]
  \centering
  \includegraphics[width=1.0\linewidth]{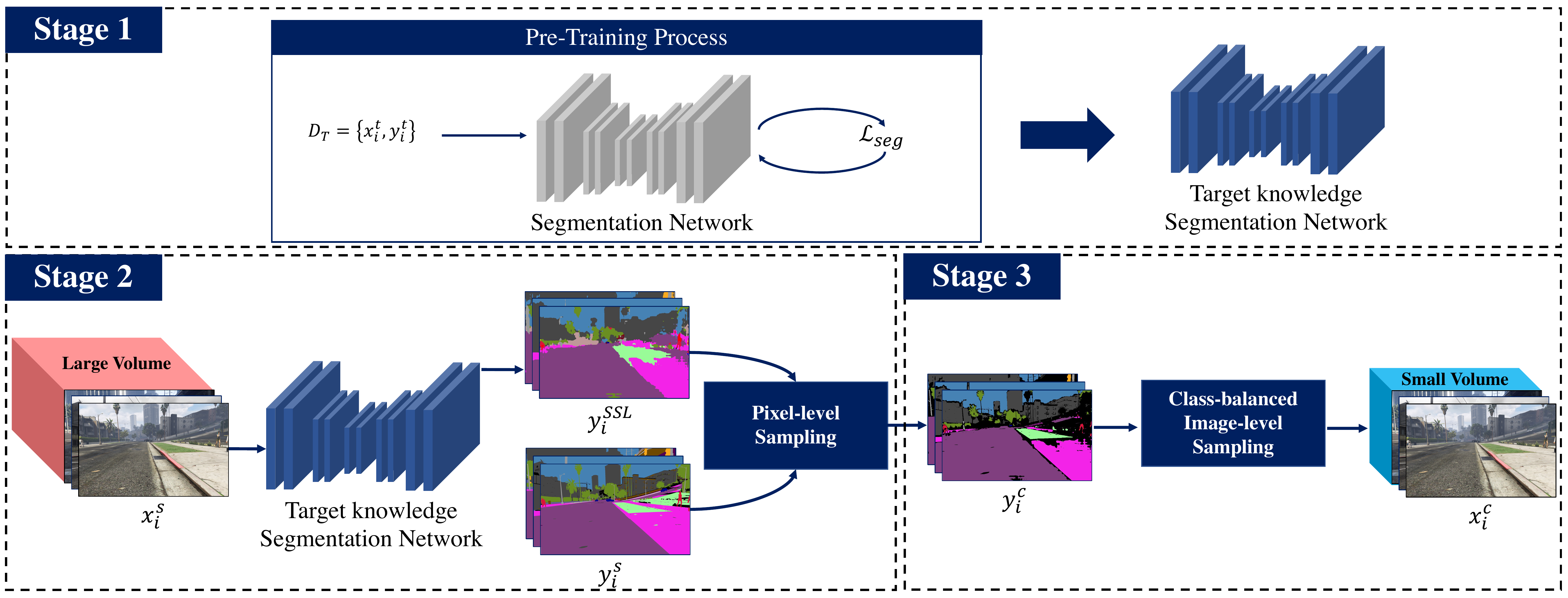}
  \caption{Pipeline of the SDSS method. It consists of a total of three stages. Stage1: Acquire knowledge of the target domain with accessible target domain data. Stage2: Generate $\mathbf{y}_{i}^{c}$ through $\mathbf{y}_{i}^{SSL}$ and $\mathbf{y}_{i}^{s}$, Stage3: Get a subset based on $\tau_{C}$ considering the class balance of $\mathbf{y}_{i}^{c}$.}
    \label{figure:figure02} 
\end{figure*}

\begin{itemize}
\setlength\itemsep{0em}
    \item We propose a novel SDSS method for semi-supervised domain adaptation.
    \item We propose a pixel-level pseudo-label sampling algorithm to exploit source samples relevant to the target domain.
    \item We propose an image-level sampling algorithm to further consider the class balance and prediction correctness of source samples.
    \item We presented a new \textit{Ocean Ship} dataset with 500 real and 200K synthetic images.
\end{itemize}

\section{Related Works}
\label{sec:related}
In this section, we describe existing DA methods and active learning methods to discuss the effect of data selection for training. In addition, we investigate how self-training methods are used in various tasks.\\

\noindent\textbf{Domain adaptation.} Among the numerous tasks of computer vision, the segmentation task is particularly time-consuming and expensive to produce labels.

To solve this problem, Richter \textit{et al.}~\cite{richter2016playing} suggested the use of synthetic data that can easily obtain labels through the use of GTA5 games.
Since then, numerous UDA studies ~\cite{zou2018unsupervised, zhang2019category, mei2020instance, wang2020differential, musto2020semantically, yang2020fda} have been proposed to improve the performance by training only with labels of synthetic data without labels of real data. Furthermore, SSDA studies~\cite{wang2020alleviating, jiang2020bidirectional, saito2019semi} have been proposed to additionally utilize a small amount of labelled target domain data together with a large amount of source domain data.
However, previous works do not pay much attention on how to efficiently utilize large-scale source data to reduce computational costs while preserving or even improving segmentation performance on the target domain.\\
\noindent\textbf{Active learning.} Because labelling a large number of data is costly and time-consuming, active learning~\cite{yoo2019learning, mayer2020adversarial, zhang2020state} has been proposed to extract samples that are helpful for further performance improvement among unlabelled data. 
Active learning first trains a model with only a small amount of labelled target data and gradually increases the number of training data by selecting sample data to be labelled from a large-scale unlabelled data for further training.
The motivation and training process of active learning is similar to our synthetic data sampling method.
However, our method has two major differences from the existing active learning: i) Active learning uses all unlabelled real data, but we only use a small amount of labelled real data. 
ii) Active learning does not use synthetic data, so there is no need to solve the domain difference problem.
However, we need to solve the domain difference problem because we sample labelled synthetic data with a small amount of labelled real data.\\
\noindent\textbf{Self-training.} The self-training method is widely used in various computer vision tasks. 
For example, various self-training methods~\cite{li2019bidirectional, zou2018unsupervised, zhang2019category, mei2020instance} have achieved state-of-the-art performance in UDA for semantic segmentation.
These methods improve their performance by creating pseudo-labels on unlabelled real data with a model trained on labelled synthetic data.
In knowledge distillation~\cite{44873, phuong2019distillation}, self-training compresses the amount of computation and parameters while maintaining the same performance.
It is also proven to be effective to deal with label noise with pseudo-labels~\cite{touvron2021training, wei2020circumventing}.\\
\noindent\textbf{Our work.} Different from previous works, we propose a source domain data sampling method that selects samples that can be beneficial to improve the target domain performance of the segmentation model.
Our sampling method benefits from the knowledge learned from a small amount of labelled target domain data.
The proposed method can be seamlessly combined with existing SSDA methods~\cite{saito2019semi,wang2020alleviating}.
As a result, with the reduced number of source domain data, we achieve a shorter training time and improved segmentation performance.

\section{Source Domain Subset Sampling}
\label{sec:method}

The proposed SDSS method follows the semi-supervised domain adaptation (SSDA) configuration.
The SSDA configuration allows access to large-scale labelled source domain data and a few labelled target domain data.
Let $\mathbf{D}_{S} = \left\{\left( \mathbf{x}_{i}^{s}, \mathbf{y}_{i}^{s} \right)\right\}_{i=1}^{N_{S}}$ and $\mathbf{D}_{T} = \left\{\left( \mathbf{x}_{i}^{t}, \mathbf{y}_{i}^{t} \right)\right\}_{i=1}^{N_{T}}$ be the labelled source and target data where $\mathbf{x}$ is an image, $\mathbf{y}$ is a corresponding GT label and $N_{S}$ and $N_{T}$ denote the number of source and target data, respectively. 
SDSS aims to extract a subset $\mathbf{D}_{C} = \left\{\left( \mathbf{x}_{i}^{c}, \mathbf{y}_{i}^{c} \right)\right\}_{i=1}^{N_{C}}$ of $\mathbf{D}_{S}$ containing source domain samples highly relevant to the target domain.
\figshortref{figure:figure02} shows the overall pipeline of the proposed method.

\begin{figure*}[!t]
  \centering
  \includegraphics[width=1.0\linewidth]{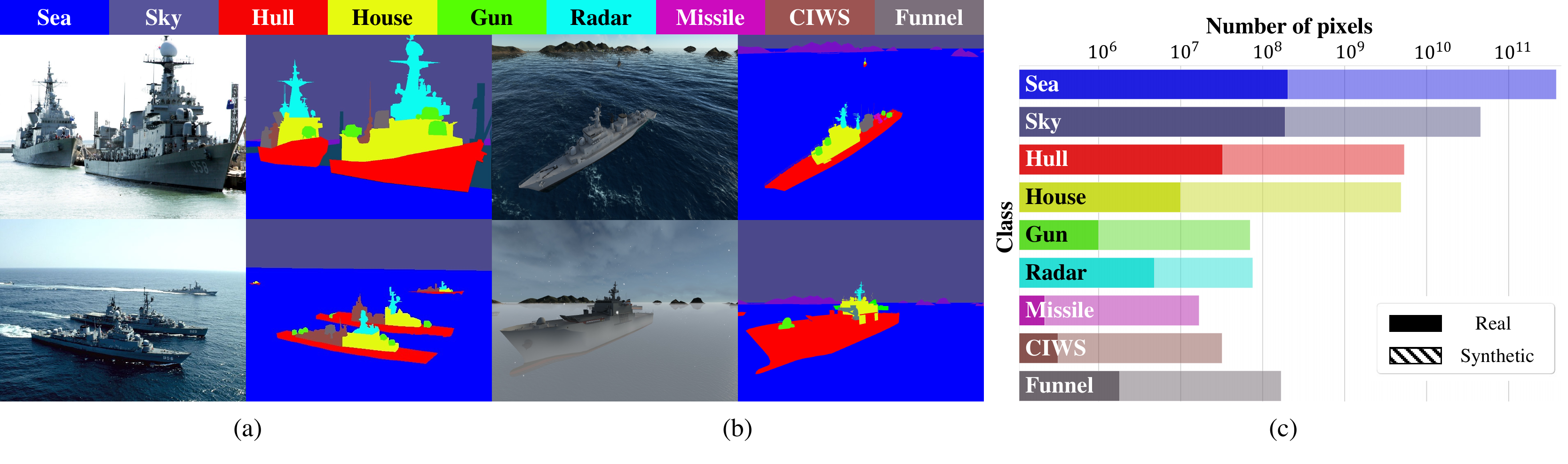}
  \caption{
  Overview of the proposed \textit{Ocean Ship} dataset. Example images of the (a) real-world and (b) synthetic data. (c) Comparison of the number of pixels in each class in synthetic and real data.\vspace{-5mm}
  }
 \label{figure:figure03}
\end{figure*}

\subsection{Pixel-level sampling}
\label{subsec:pixel_level}

One of the main reasons for performance degradation in DA is an inherent domain gap between the source and target domains.
To minimize this domain gap, we refine source domain data in pixel-level based on the knowledge learned from the available small-scale target domain data.
Therefore, we extract a pre-trained model biased to the target domain by training on a small amount of labelled data, as shown in stage 1 in~\figshortref{figure:figure02}.
Then, as shown in stage 2 in ~\figshortref{figure:figure02}, all source domains are predicted with the pre-trained model, and all, except the pixel information matching the GT of the source domain, is removed.
Through this process, we can use the information of the source domain biased toward the target domain at the pixel-level.

Following \cite{li2019bidirectional}, we first generated a pseudo-label $\mathbf{y}_{i}^{SSL}$ of the pixel $i$ from the source domain for self-supervision as follows:

\begin{equation}
\scalemath{0.8}{
    \mathbf{y}_{i}^{SSL} = \begin{cases}
        \underset{k}{\argmax}~\mathbf{M}\left(\mathbf{x}_{i}^{s}\right)_{k} & \mathrm{if}~~ \mathbf{M}\left(\mathbf{x}_{i}^{s}\right)_{k} \geq \tau_{SSL} ~~\mathrm{where}~~ k \in [1, K], \\
        -1, & \mathrm{otherwise},
    \end{cases}}
\label{eq:pseudo_label}
\end{equation}

where $\mathbf{M}(\cdot)$ is a segmentation network and $k$, $K$, and $\tau_{SSL}$ denote the class index, number of classes, and prediction confidence threshold set to 0.1.

Next, we further refined $\mathbf{y}_{i}^{SSL}$ by comparing it with the corresponding GT $\mathbf{y}_{i}^{s}$ to ensure the reliability of the generated pseudo-labels.
In other words, pseudo-labels that are not matched with the GT labels were eliminated to obtain our refined pseudo-labels $\mathbf{y}_{i}^{c}$ as follows:

\begin{equation}
    \mathbf{y}_{i}^{c} = \begin{cases}
        \mathbf{y}_{i}^{SSL} & \mathrm{if}~~ \mathbf{y}_{i}^{SSL} = \mathbf{y}_{i}^{s}, \\
        -1, & \mathrm{otherwise}.
    \end{cases}
\label{eq:refined_pseudo_label}
\end{equation}

\subsection{Image-level sampling}
\label{subsec:image_level}

Because our SDSS extracts a subset of the source domain with a small amount of available target domain information, the performance of our SDSS varies according to the class balance information of the initially trained target domain. To solve this problem, we propose an image-level sampling based on class balance scoring that extracts a subset of the source domain considering the class balance of the target domain. The class balance score is defined as follows:

\begin{equation}
    \tau_{C} = \sum_{k=1}^{K}\frac{n_{k}^{correct}}{n_{k}^{class}} \cdot \left(1 - \frac{n_{k}^{class}}{n^{image}} \right),
\label{eqn:image_score}
\end{equation}

where $n_{k}^{correct}$ and $n_{k}^{class}$ are the number of correct pseudo-label and GT pixels of the class $k$ in an image, respectively, and $n^{image}$ denotes the number of pixels in an image.
Our image scoring function in \eqnref{eqn:image_score} considers two aspects:
i) correctness ratio of each class, and ii) pixel ratio of each class in an image.
The larger the ratio of the number of correctly classified pixels to that of the corresponding GT ones is (i.e, correctness ratio), the higher the image score is given.
It is further adjusted based on the proportion of each class in the image where classes with a large portion of pixels will be given lower scores.
With our method, an image containing a variety of classes and correct samples has a high chance to be utilized, whereas an image with imbalanced classes or a few number of available samples can be excluded easily for the training.

\section{Experimental results}
\label{sec:experiment}

In this section, we present experimental results to validate the proposed sampling method for semantic segmentation.
We first describe experimental configurations in detail.
Then, we validate our SDSS on two public benchmark datasets, GTA5~\cite{richter2016playing} and SYNTHIA~\cite{ros2016synthia}, and provide detailed analyses.
Then, we verify and analyse the performance of our method on our new dataset called, \textit{Ocean Ship}.
Note that the Intersection-over-Union (IoU) metric was used for all the experiments.

\begin{table*}[!t]
    \caption{\label{table:table1}Quantitative evaluation results on (a) GTA5 $\rightarrow$ Cityscapes (19 classes) and (b) SYNTHIA $\rightarrow$ Cityscapes (13 classes) scenarios. Each method is evaluated on the Cityscapes validation set.\vspace{-3mm}}
\resizebox{.99\textwidth}{!}{
    \centering
    \def\arraystretch{1.2}
    \begin{tabular}{cc|cccccc|cccccc}
    \hline \hline
    \multicolumn{2}{c|}{} & \multicolumn{6}{c|}{(a) \textbf{GTA5} $\rightarrow$ \textbf{Cityscapes}} & \multicolumn{6}{c}{(b) \textbf{SYNTHIA} $\rightarrow$ \textbf{Cityscapes}} \\ \hline
    \multirow{2}{*}{Type} & \multirow{2}{*}{Method} & \multicolumn{6}{c|}{Number of labelled targets} & \multicolumn{6}{c}{Number of labelled targets} \\
    & & 0 & 100 & 200 & 500 & 1,000 & 2,975 & 0 & 100 & 200 & 500 & 1,000 & 2,975 \\ \hline
    \multirow{5}{*}{UDA} & AdaptSeg~\cite{tsai2018learning} & 42.4 & - & - & - & - & - & 46.7 & - & - & - & - & - \\
    & Advent~\cite{vu2019advent} & 45.5 & - & - & - & - & - & 48.0 & - & - & - & - & - \\
    & BDL~\cite{li2019bidirectional} & 48.5 & - & - & - & - & - & 51.4 & - & - & - & - & - \\
    & CADASS~\cite{yang2021context} & 49.2 & - & - & - & - & - & 52.4 & - & - & - & - & - \\ 
    & FDA~\cite{yang2020fda} & 50.45 & - & - & - & - & - & 52.5 & - & - & - & - & - \\ \hline
    Supervised & Deeplab-v2 & - & 41.70 & 48.16 & 53.29 & 57.50 & 63.03 & - & 54.56 & 58.27 & 62.39 & 65.64 & 69.86 \\ \hline
    \multirow{7}{*}{SSDA} & Baseline & - & 46.10 & 50.29 & 53.52 & 55.97 & 59.78 & - & 55.35 & 56.84 & 60.96 & 63.22 & 67.80 \\
    & \cellcolor{Gray} Baseline + Ours & \cellcolor{Gray} - & 
    \cellcolor{Gray} \textbf{50.37(+4.27)} & 
    \cellcolor{Gray} \textbf{51.57(+1.28)} & 
    \cellcolor{Gray} \textbf{55.62(+2.10)} & 
    \cellcolor{Gray} \textbf{58.53(+2.56)} & 
    \cellcolor{Gray} \textbf{63.48(+2.30)} & 
    \cellcolor{Gray} - & 
    \cellcolor{Gray} \textbf{57.80(+2.45)} & 
    \cellcolor{Gray} \textbf{60.42(+3.58)} & 
    \cellcolor{Gray} \textbf{63.19(+2.23)} & 
    \cellcolor{Gray} \textbf{65.78(+2.56)} & 
    \cellcolor{Gray} \textbf{70.04(+2.24)} \\
    & MME~\cite{saito2019semi} & - & 51.42 & 53.46 & 56.70 & 60.57 & 63.11 & - & 58.04 & 61.96 & 65.70 & 67.36 & 69.90 \\
    & \cellcolor{Gray} MME + Ours & \cellcolor{Gray} - & 
    \cellcolor{Gray} \textbf{54.08(+2.66)} & 
    \cellcolor{Gray} \textbf{56.16(+2.70)} & 
    \cellcolor{Gray} \textbf{58.61(+1.91)} & 
    \cellcolor{Gray} \textbf{62.17(+1.60)} & 
    \cellcolor{Gray} \textbf{64.88(+1.77)} & 
    \cellcolor{Gray} - & 
    \cellcolor{Gray} \textbf{59.22(+1.18)} & 
    \cellcolor{Gray} \textbf{63.43(+1.47)} & 
    \cellcolor{Gray} \textbf{67.27(+1.57)} & 
    \cellcolor{Gray} \textbf{68.40(+1.04)} & 
    \cellcolor{Gray} \textbf{70.91(+1.01)} \\
    & ASS~\cite{wang2020alleviating} & - & 53.15 & 55.21 & 60.08 & 63.12 & 68.64 & - & 61.74 & 63.80 & 68.65 & 72.40 & 75.77 \\
    & \cellcolor{Gray} ASS + Ours & 
    \cellcolor{Gray} - &
    \cellcolor{Gray} \textbf{56.14(+2.99)} & \cellcolor{Gray} \textbf{58.42(+3.21)} & \cellcolor{Gray} \textbf{62.84(+2.76)} & \cellcolor{Gray} \textbf{65.68(+2.56)} & \cellcolor{Gray} \textbf{69.65(+1.01)} & \cellcolor{Gray} - & \cellcolor{Gray} \textbf{64.20(+2.46)} &
    \cellcolor{Gray}\textbf{66.44(+2.64)} & \cellcolor{Gray} \textbf{69.91(+1.26)} & \cellcolor{Gray} \textbf{73.51(+1.11)} & \cellcolor{Gray} \textbf{77.54(+1.77)} \\
    \hline
    \multicolumn{2}{c|}{Number of labelled sources(Ours)} & & 17,481 & 16,622 & 19,695 &20,039&20,764& &3,333&3,478&4,409&4,244&4,533\\
    \hline \hline
    \end{tabular}}
\end{table*}

\subsection{Dataset}
We evaluated our SDSS on two popular semantic segmentation DA benchmarks, GTA5 $\rightarrow$ Cityscapes and SYNTHIA $\rightarrow$ Cityscapes, and our new \textit{Ocean Ship}(synthetic $\rightarrow$ real).

\noindent\textbf{Cityscapes.} 
Cityscapes is a real-world urban scene dataset consisting of 2,975 training, 500 validation, and 1,525 testing images.
Following the standard protocols~\cite{saito2019semi, wang2020alleviating}, we randomly selected 100, 200, 500, 1,000 and 2,975 images from the training images sequentially for training, and evaluated our SDSS method on the 500 validation images.

\noindent\textbf{GTA5.} The GTA5 dataset is a synthetic dataset sharing 19 semantic classes with Cityscapes.
A total of 24,966 urban scene images were collected from the video game \textit{Grand Theft Auto V}, and were used as the source domain training data.

\noindent\textbf{SYNTHIA.} SYNTHIA is a synthetic urban scene dataset. We utilized SYNTHIA-RAND-CITYSCAPES which shares 16 semantic classes with Cityscapes as the source domain dataset.
In total, 9,400 images from the SYNTHIA dataset were used as source domain training data.

\noindent\textbf{\textit{Ocean Ship}.}
Combining large-scale synthetic data with small-scale real-world ones for deep network training is an effective configuration to tackle the lack of large-scale real-world data~\cite{roitberg2021let}.
Unfortunately, the existing ocean scenario datasets ~\cite{shao2018seaships, moosbauer2019benchmark} are not large-scale compared to other scenarios~\cite{huang2019apolloscape, yu2020bdd100k, geiger2012we} relatively. One of the main reasons is the difficulty of real-world data collection.
However, large-scale ocean scenario datasets are important because they can be used for various practical applications such as coastal surveillance.
Therefore, we have constructed the \textit{Ocean Ship} dataset consisting of 500 real ocean ship images collected from the Internet and 200K synthetic images with pixel-level annotations created using Unity-based simulation tool.
Each part of a ship is categorized into 9 classes: \{sea, sky, hull, house, gun, radar, missile, CIWS, and funnel\}.
\figshortref{figure:figure03} shows the sample images and class distribution of the \textit{Ocean Ship} dataset.
Our dataset provides images with various viewpoints and weather changes similar to those of existing datasets~\cite{richter2016playing, ros2016synthia, cordts2016cityscapes}.
%

\subsection{Implementation details}

Following previous works~\cite{zou2018unsupervised, zhang2019category, mei2020instance}, we adopted Deeplab-v2~\cite{chen2017deeplab} with the pre-trained ResNet101~\cite{he2016deep} on ImageNet~\cite{deng2009imagenet} as our network architecture.
We also followed the training configuration for a fair comparison. Accordingly, we set the batch size to 1, learning rate to $2.5\times10^{-4}$, and the number of iteration to 250,000 and early stopped at 120,000 iterations.
To verify the effectiveness of our SDSS, we trained the network with a small number of labelled target domain data $\mathbf{D}_{T}$ and source domain data $\mathbf{D}_{S}$ as $\textit{Baseline}$.
Then, the network was trained with $\mathbf{D}_{T}$ and the subsampled source domain data $\mathbf{D}_{C}$.
To verify the sampling effect in a fair manner, the target knowledge network used for source domain data sampling is not used for the subsequent training steps.

\subsection{Quantitative evaluation on semantic segmentation datasets}

In this section, we compare the performance of our SDSS method with state-of-the-art UDA and SSDA methods on public benchmark datasets. In addition, we verify the effectiveness of the SDSS on our new \textit{Ocean Ship} dataset. 

\noindent\textbf{GTA5 to Cityscapes.} \tabref{table:table1}{(a)} shows the performance comparison of UDA, fully-supervised, and SSDA methods in the GTA5 to Cityscapes scenario.
UDA methods~\cite{tsai2018learning,vu2019advent,li2019bidirectional,yang2020fda} show promising results with only tens of thousands of source domain data without any target domain data.
However, if a few labelled target domain data are available, then the fully-supervised method shows a much better performance.

In SSDA, if the source and target domain data are directly combined for the training, then it is helpful only when the target domain labelled data are lacking.
By contrast, it becomes harmful when large-scale target domain GT are provided.
However, with our source domain subsampling method, performance improvement is consistently ensured with additional source domain data regardless of the number of the available target domain data.
The proposed method (i.e, SSDA Baseline + Ours) always outperforms the fully-supervised model (i.e, Supervised Deeplab-v2).
Furthermore, our method can be seamlessly combined with existing SSDA methods~\cite{saito2019semi,wang2020alleviating} for further performance improvement.
We conclude that the role of the proposed SDSS is complementary to the existing SSDA methods.

\begin{figure*}[!t]
\resizebox{0.95\textwidth}{!}{
\centering
    \def\arraystretch{0.1}
    \begin{tabular}{@{\hskip 0.0015\linewidth}c@{\hskip 0.0015\linewidth}c@{\hskip 0.0015\linewidth}c@{\hskip 0.0015\linewidth}c@{\hskip 0.0015\linewidth}c@{\hskip 0.0015\linewidth}c@{\hskip 0.0015\linewidth}c}
    \multicolumn{6}{c}{\scriptsize \textbf{(a) GTA5} $\rightarrow$ \textbf{Cityscapes}} & \\
    \vspace{+0.05mm}& \vspace{+0.05mm}& \vspace{+0.05mm}& \vspace{+0.05mm}& \vspace{+0.05mm}& \vspace{+0.05mm}& \\
    \includegraphics[width=0.15\linewidth]{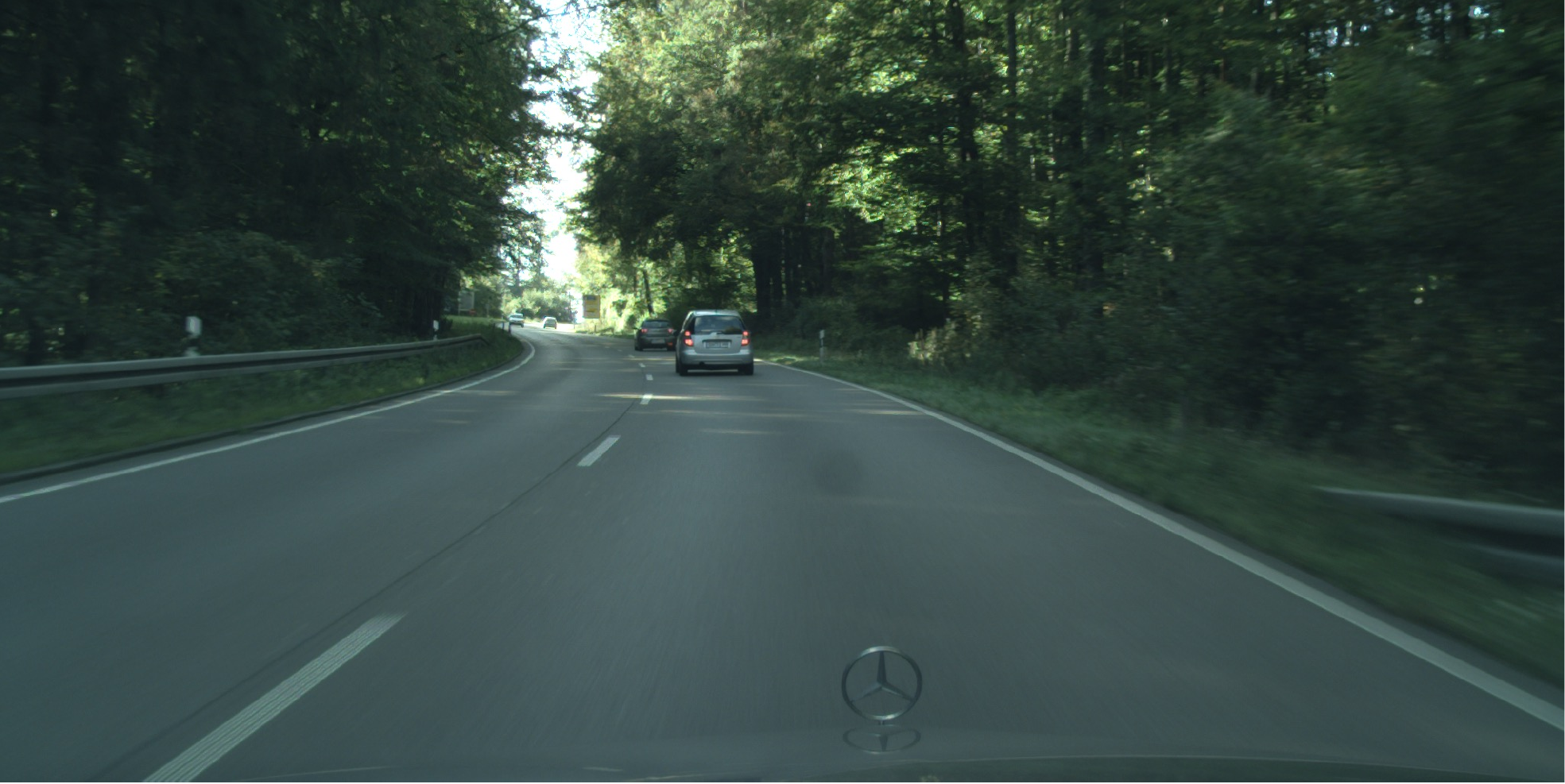} & \includegraphics[width=0.15\linewidth]{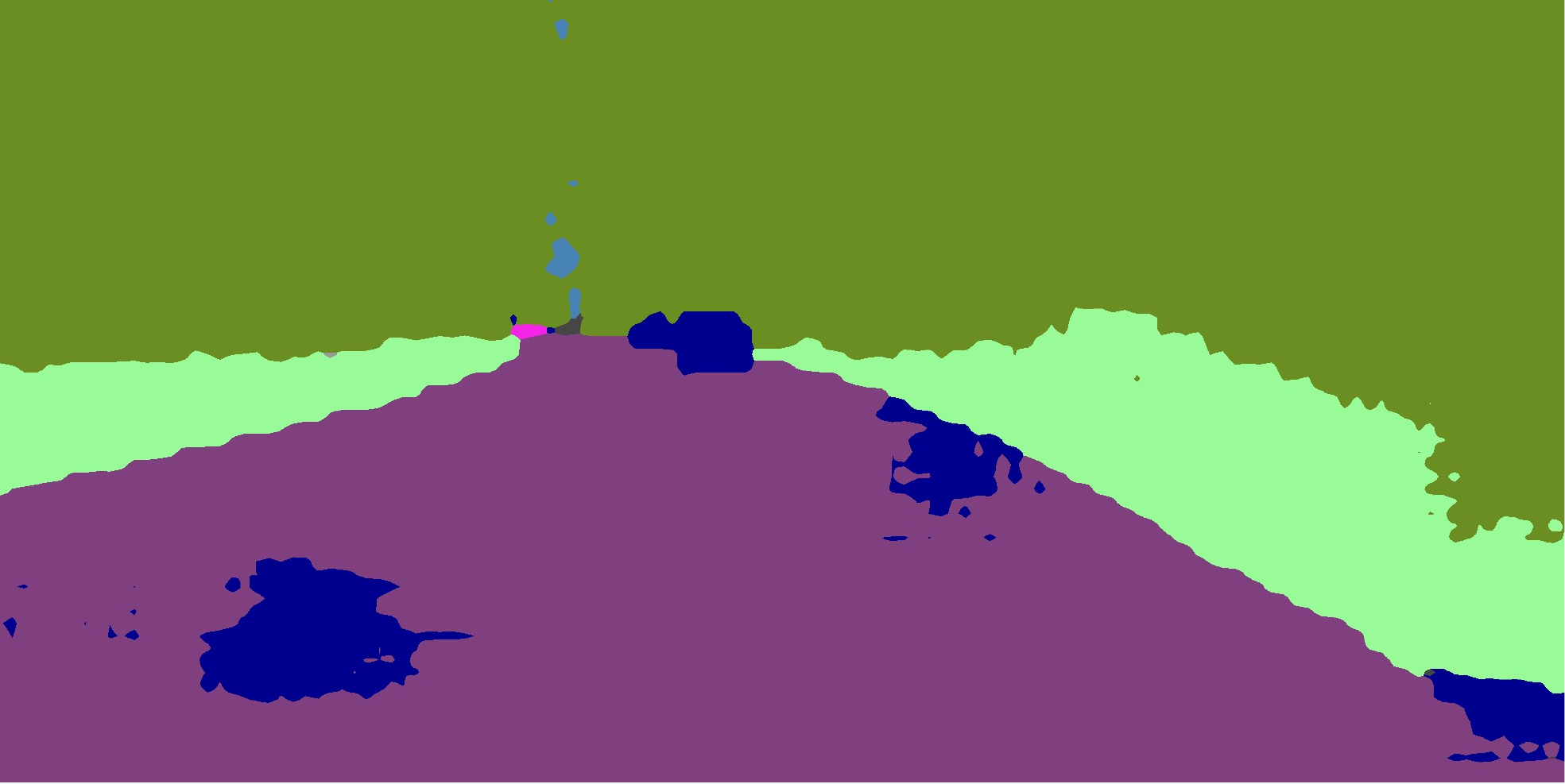} & \includegraphics[width=0.15\linewidth]{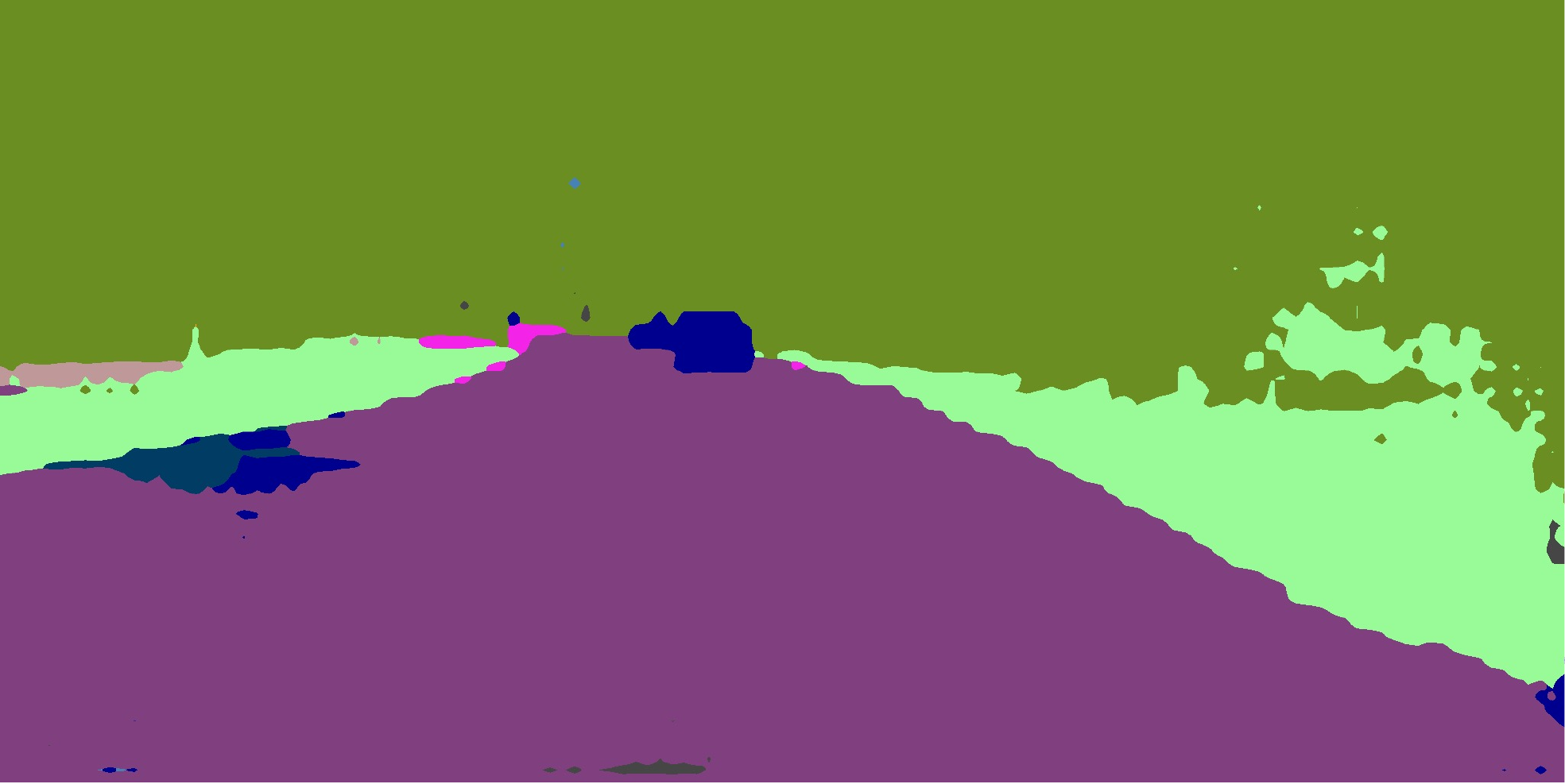} &
    \includegraphics[width=0.15\linewidth]{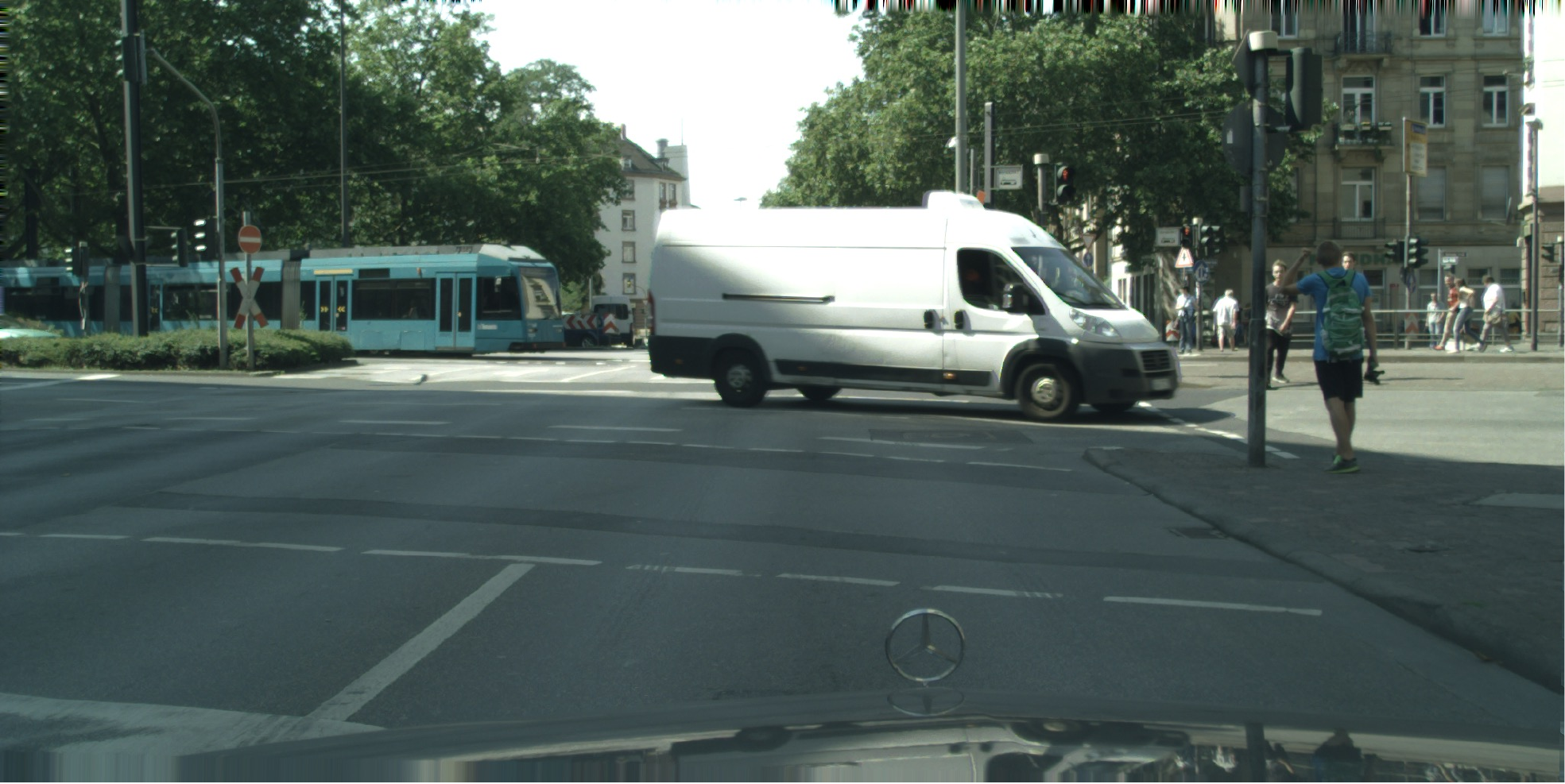} &
    \includegraphics[width=0.15\linewidth]{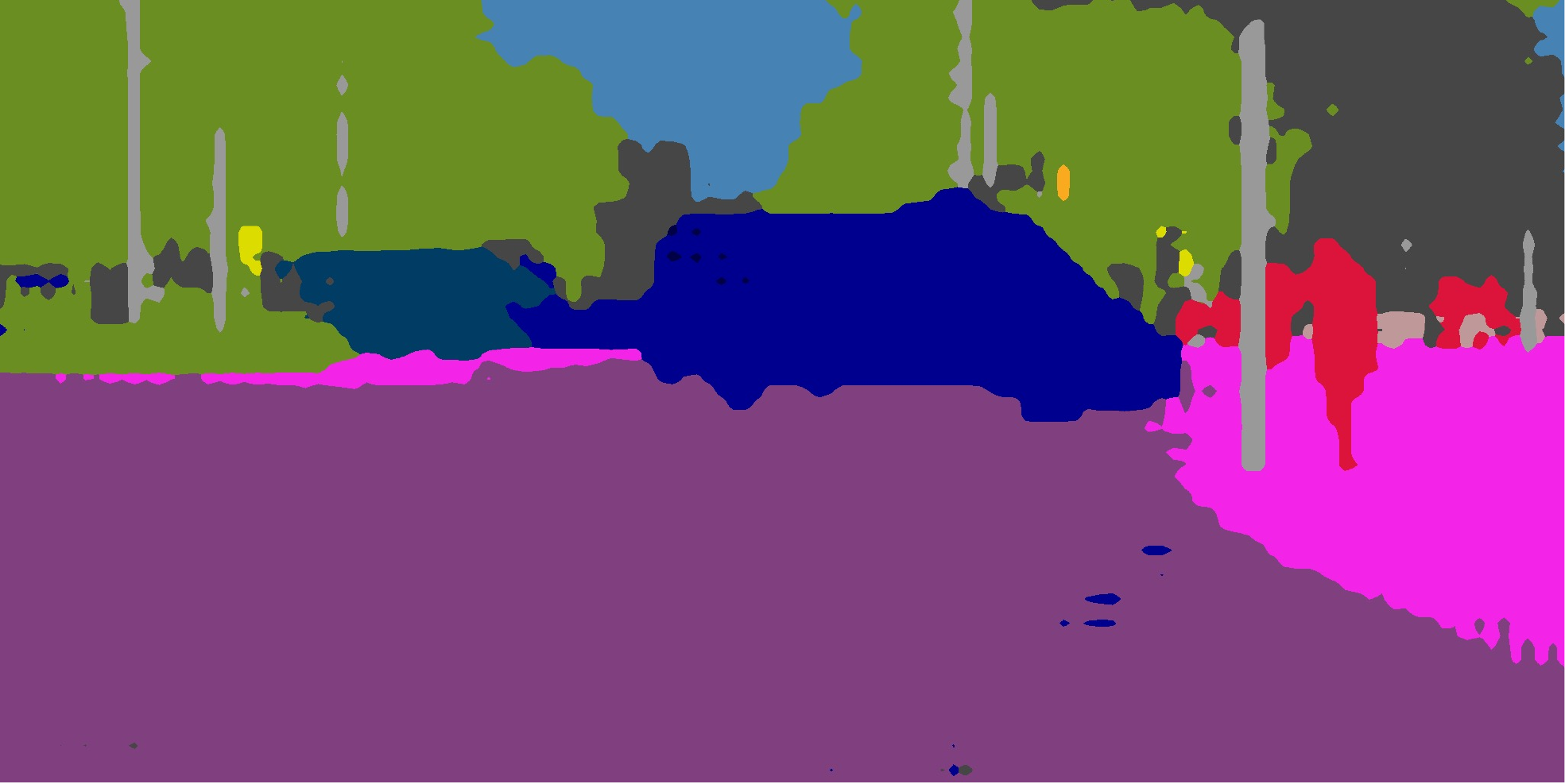} &
    \includegraphics[width=0.15\linewidth]{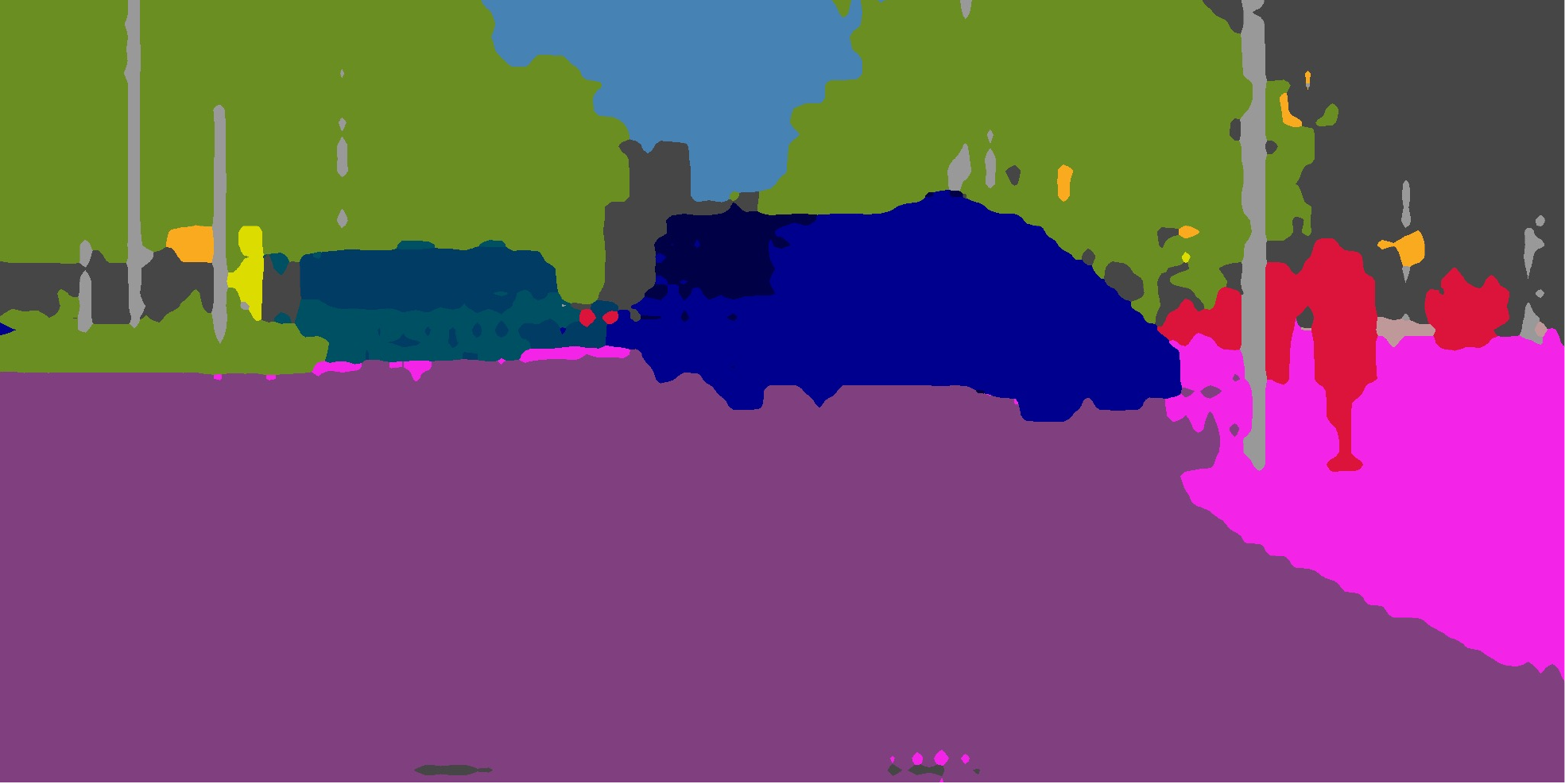} & \hspace{+0.5mm} \scriptsize\rotatebox{90}{\,\,\,\,\,Ours}\\

    \includegraphics[width=0.15\linewidth]{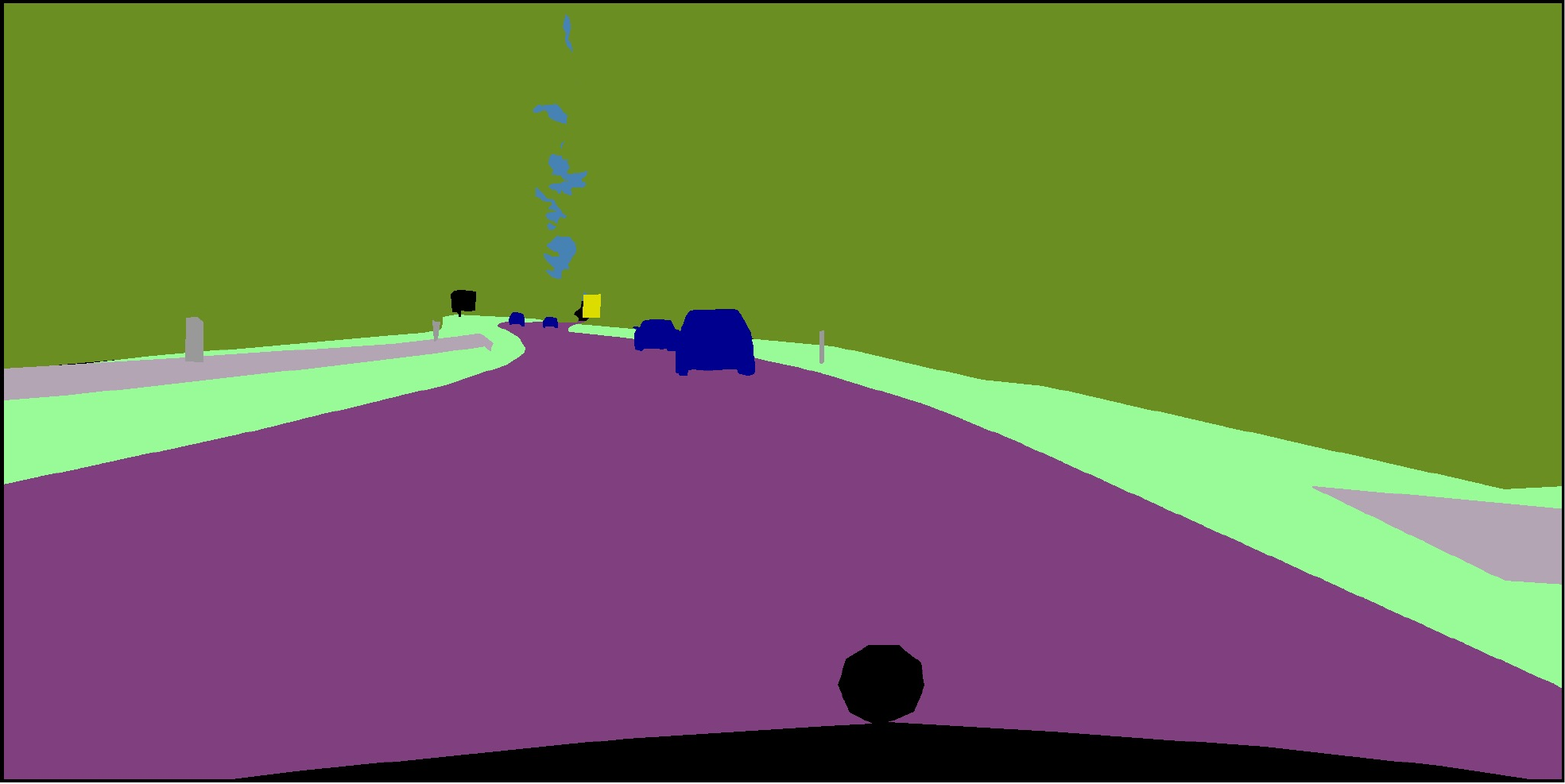} & \includegraphics[width=0.15\linewidth]{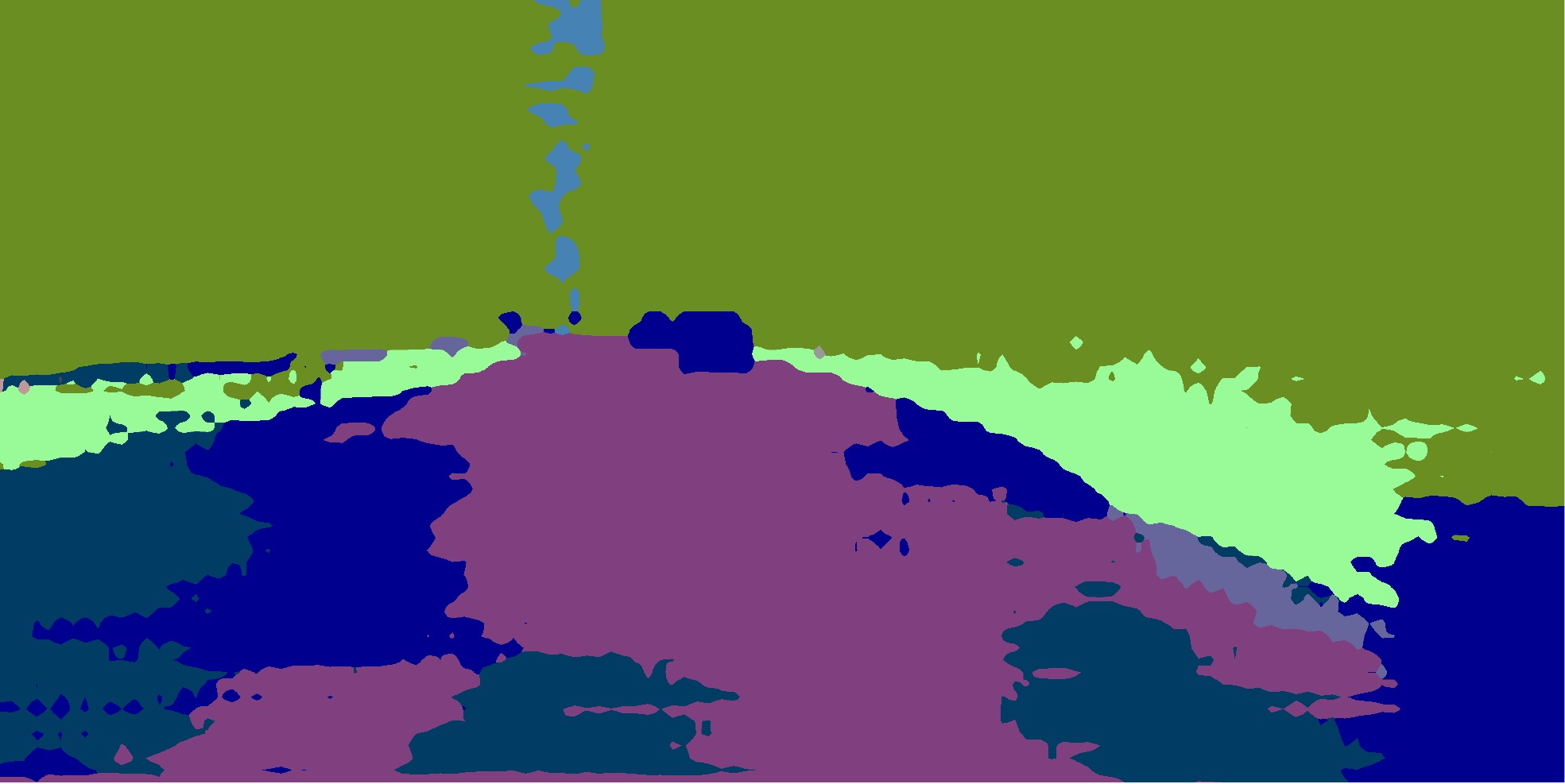} & \includegraphics[width=0.15\linewidth]{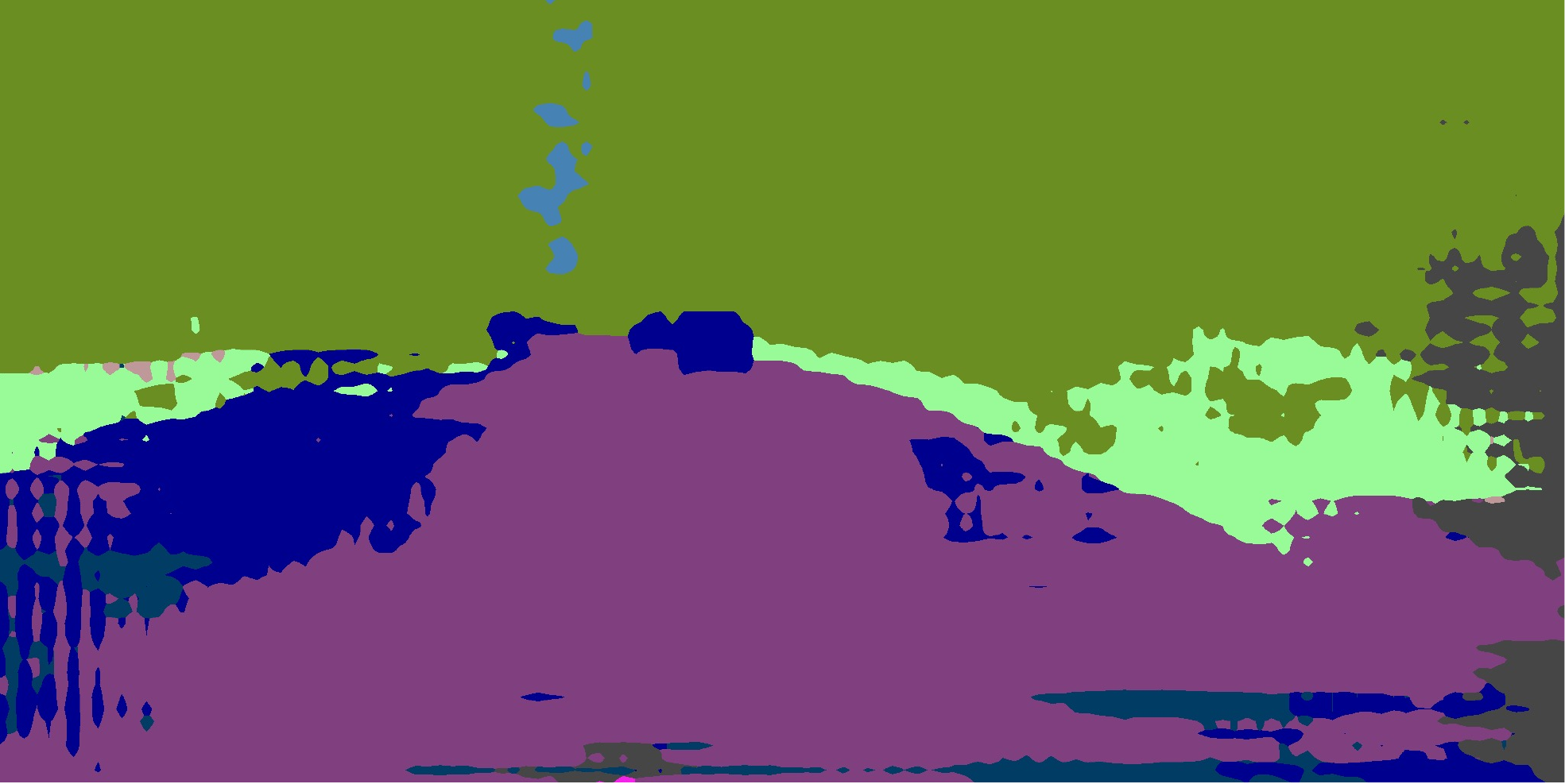} &
    \includegraphics[width=0.15\linewidth]{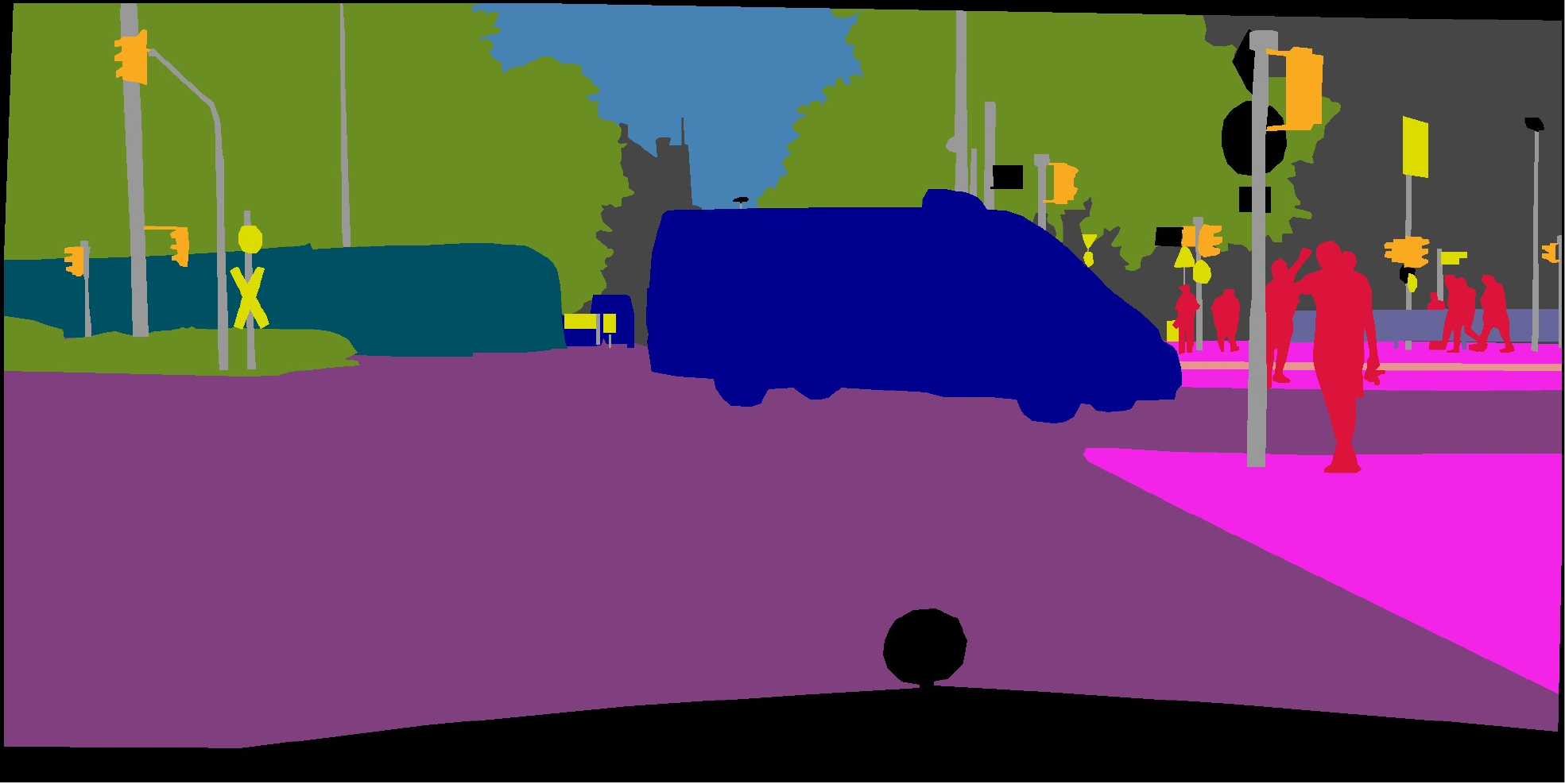} &
    \includegraphics[width=0.15\linewidth]{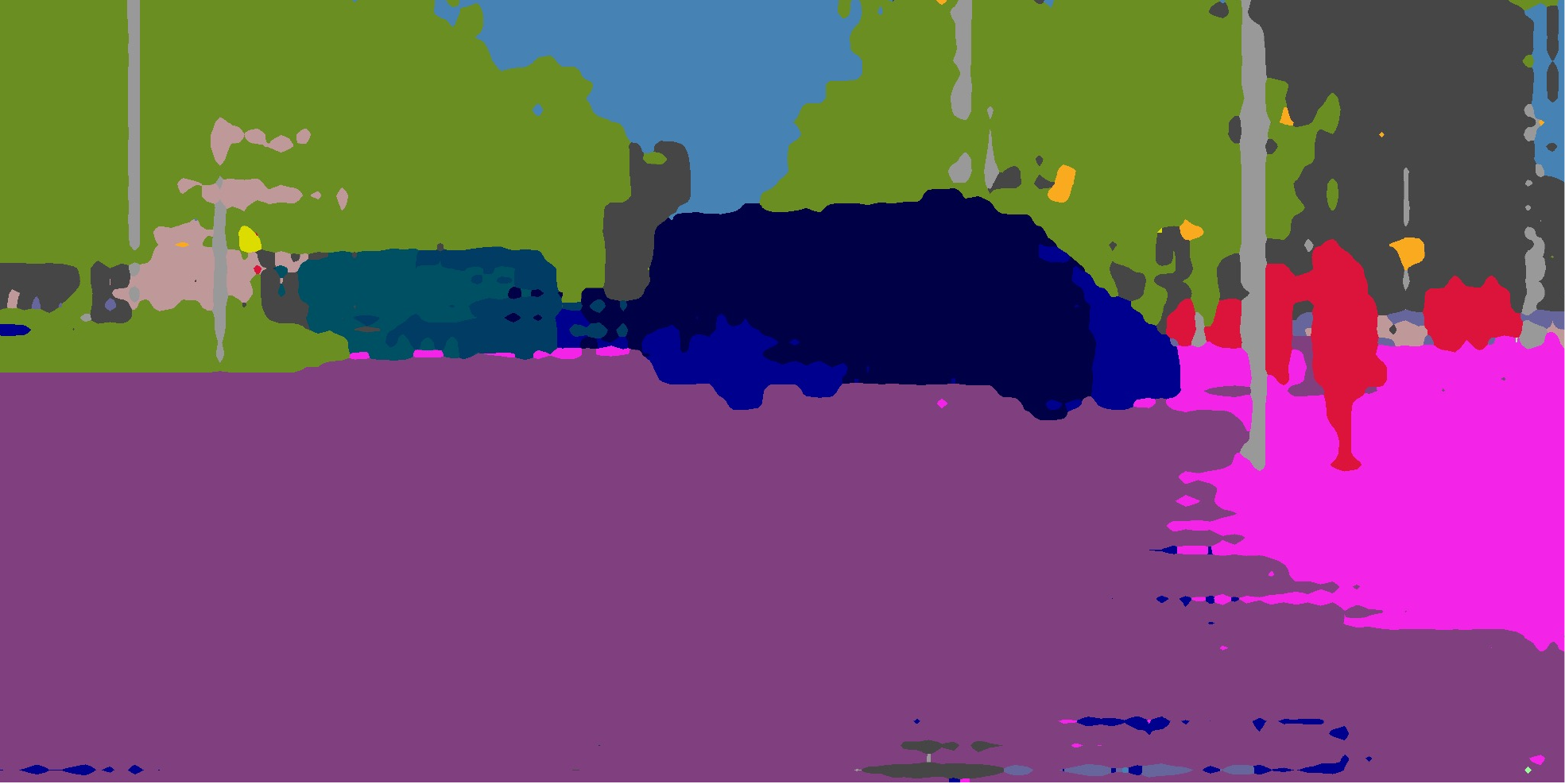} &
    \includegraphics[width=0.15\linewidth]{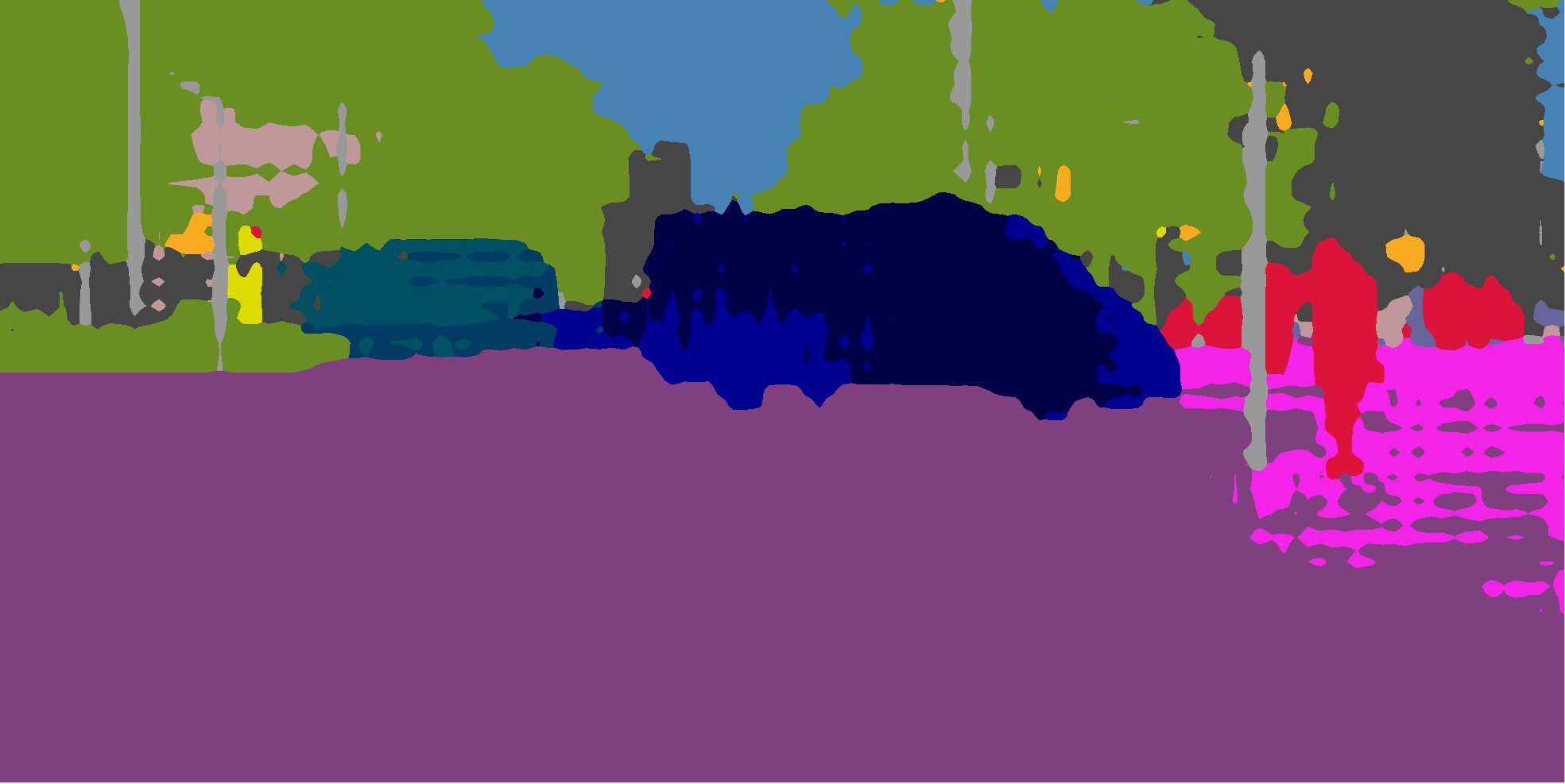} & \hspace{+0.5mm} \scriptsize\rotatebox{90}{\,Baseline} \\
    \vspace{+0.15mm}& \vspace{+0.15mm}& \vspace{+0.15mm}& \vspace{+0.15mm}& \vspace{+0.15mm}& \vspace{+0.15mm}& \\
    \multicolumn{3}{c}{\scriptsize \textbf{(b) SYNTHIA} $\rightarrow$ \textbf{Cityscapes}} &   \multicolumn{3}{c}{\scriptsize \textbf{(c) \textit{Ocean Ship} (synthetic $\rightarrow$ real)}}& \\
    \vspace{+0.05mm}& \vspace{+0.05mm}& \vspace{+0.05mm}& \vspace{+0.05mm}& \vspace{+0.05mm}& \vspace{+0.05mm}& \\
    \includegraphics[width=0.15\linewidth]{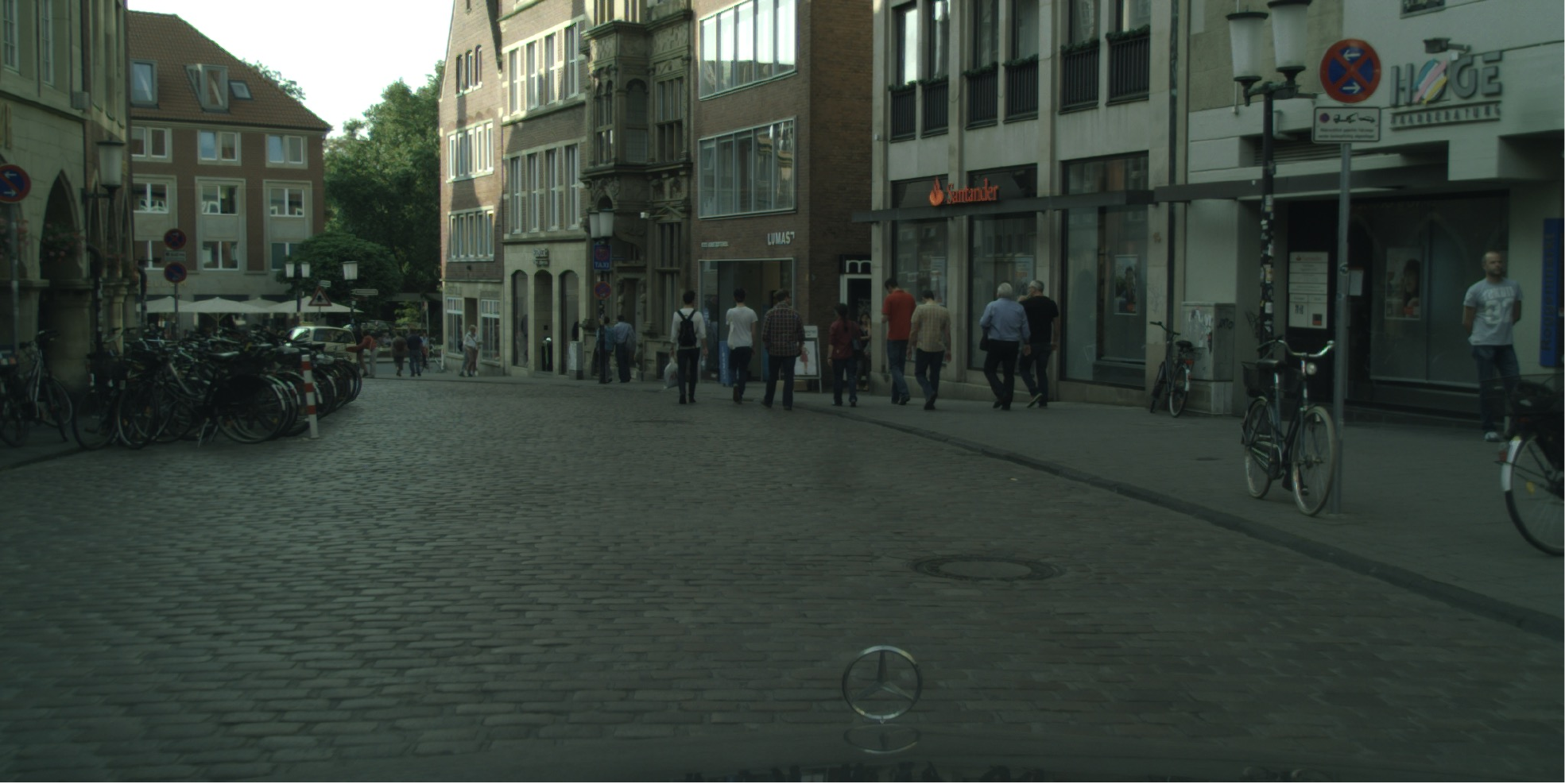} & \includegraphics[width=0.15\linewidth]{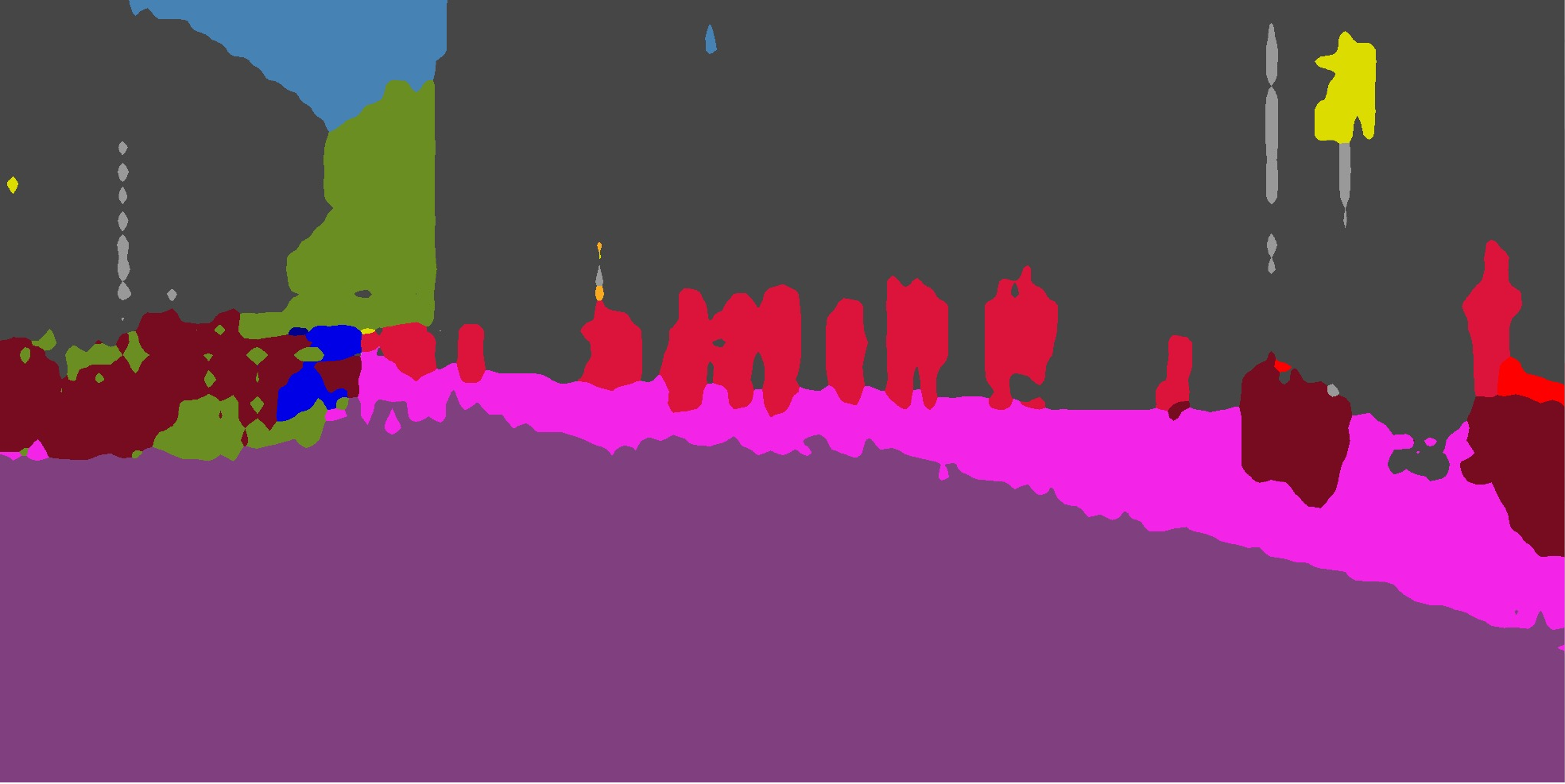} & \includegraphics[width=0.15\linewidth]{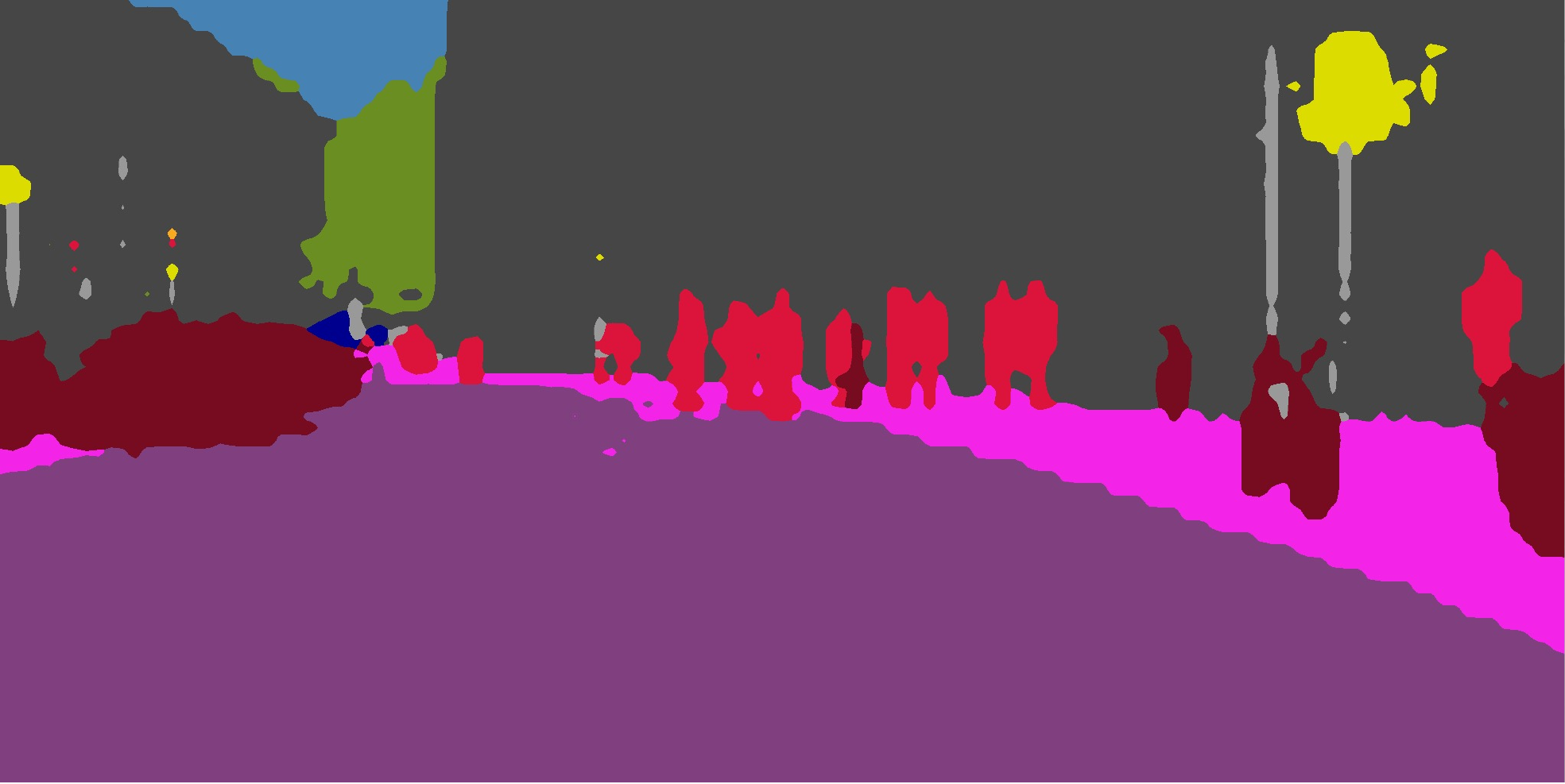} &
    \includegraphics[width=0.15\linewidth]{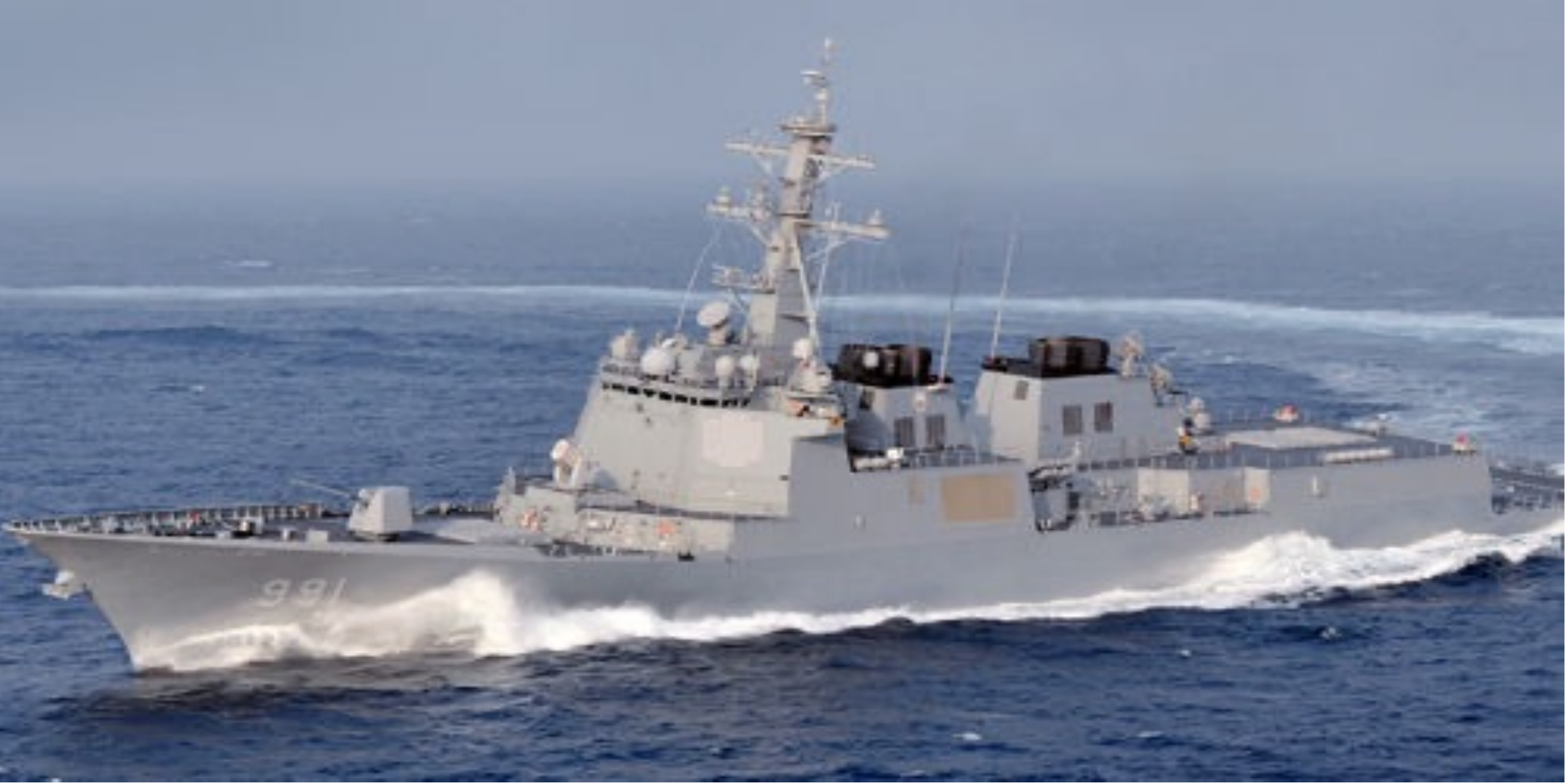} &
    \includegraphics[width=0.15\linewidth]{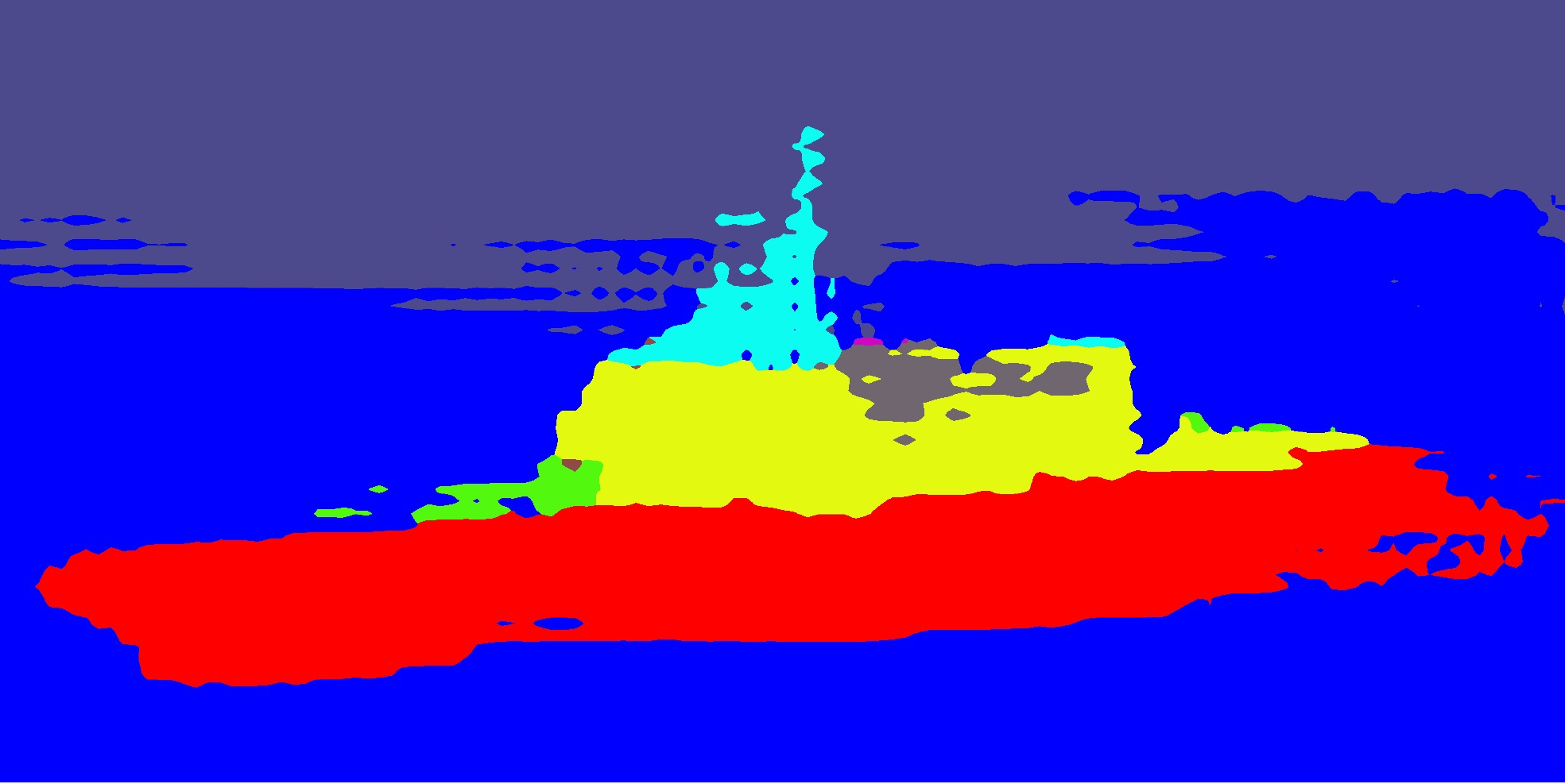} &
    \includegraphics[width=0.15\linewidth]{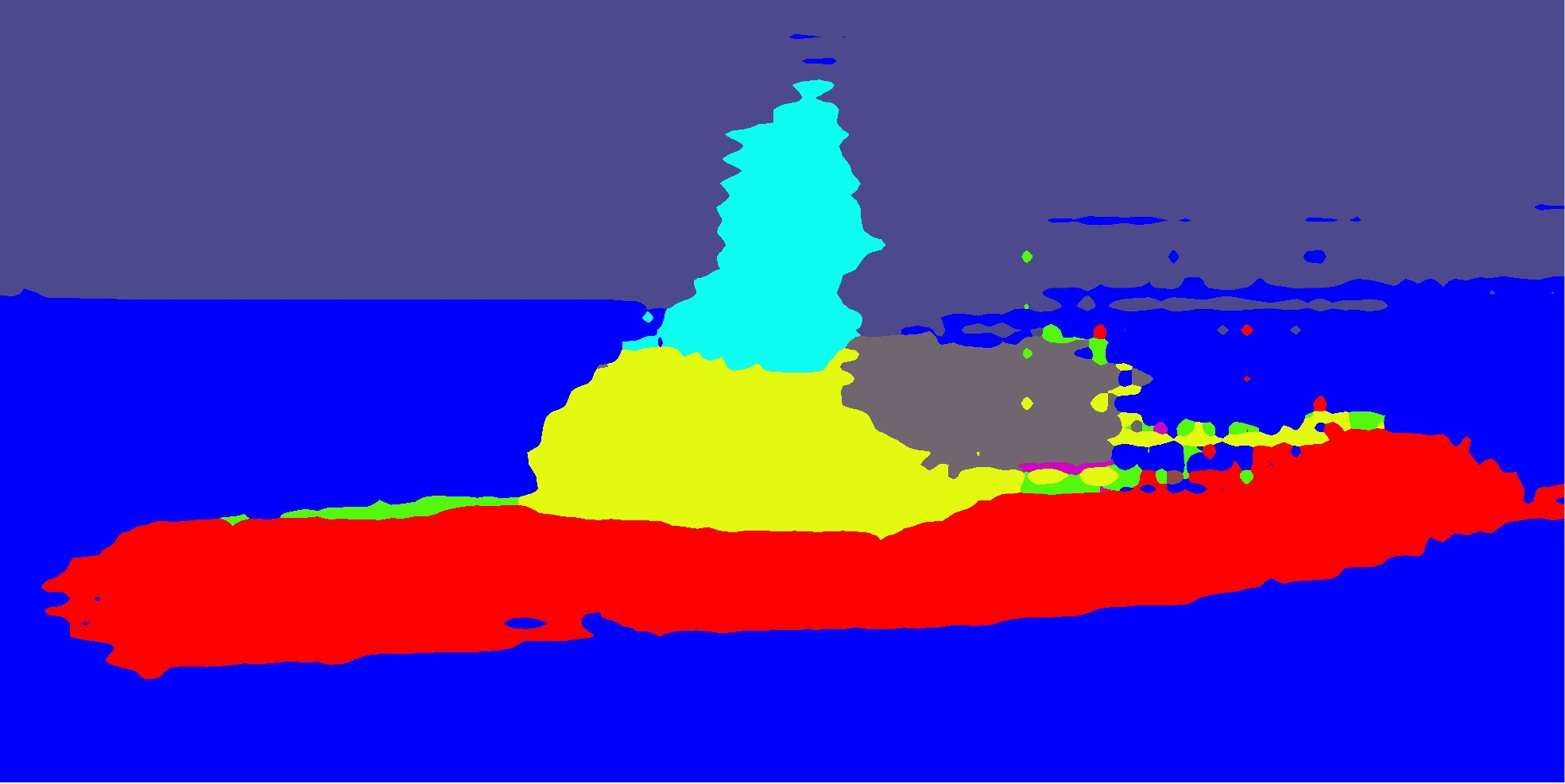} & \hspace{+0.5mm} \scriptsize\rotatebox{90}{\,\,\,\,\,Ours}\\
    \includegraphics[width=0.15\linewidth]{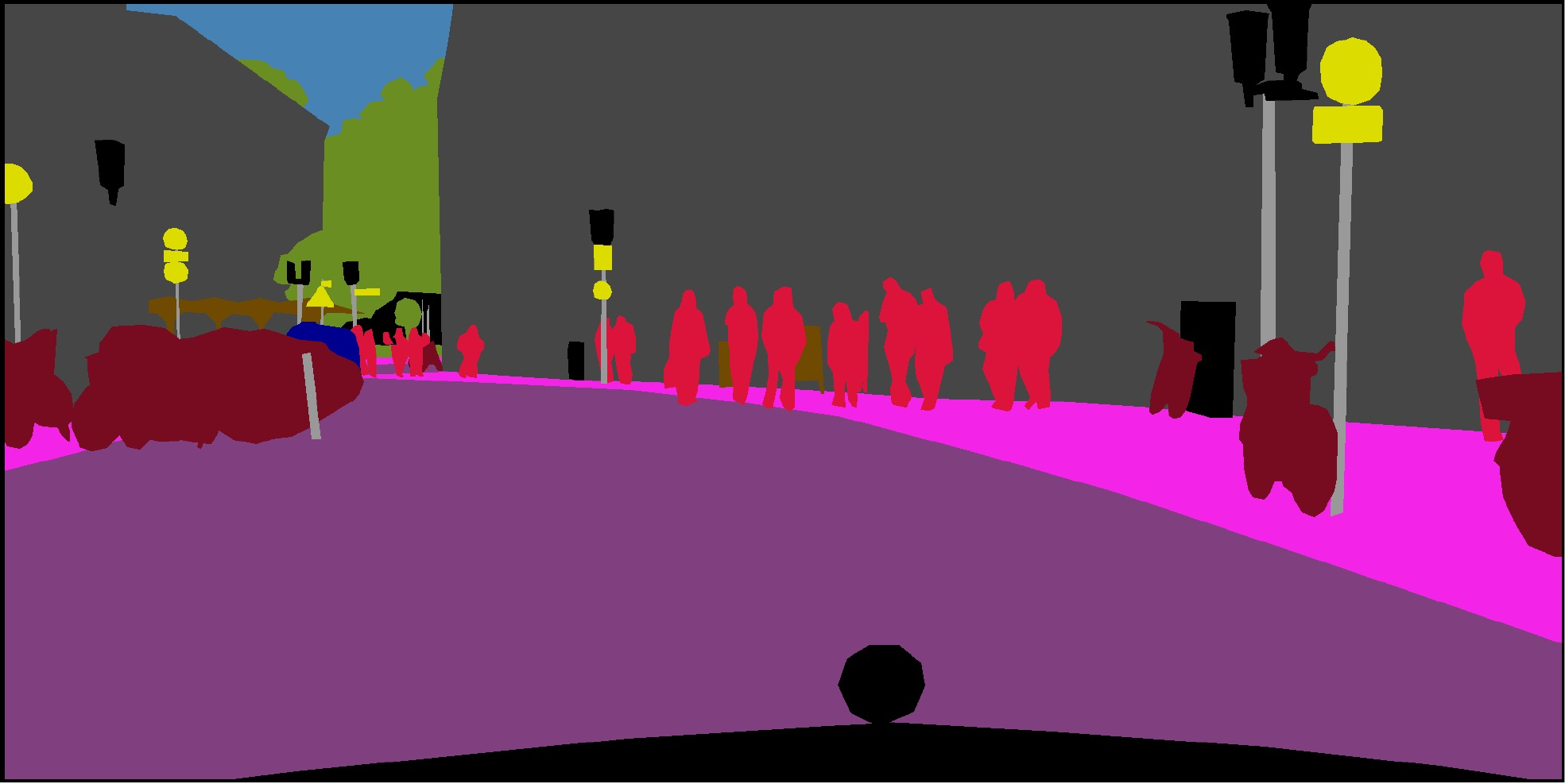} & \includegraphics[width=0.15\linewidth]{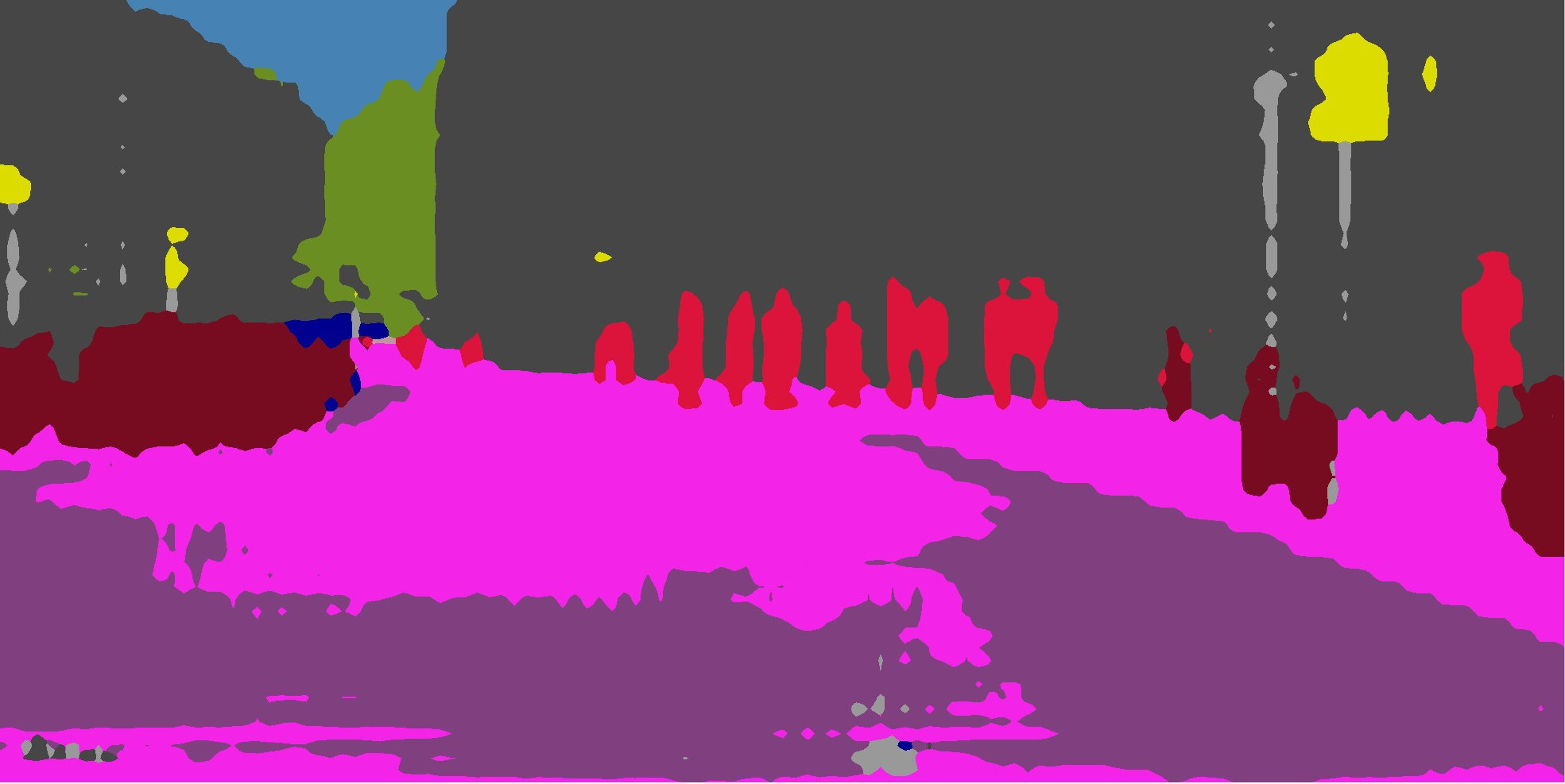} & \includegraphics[width=0.15\linewidth]{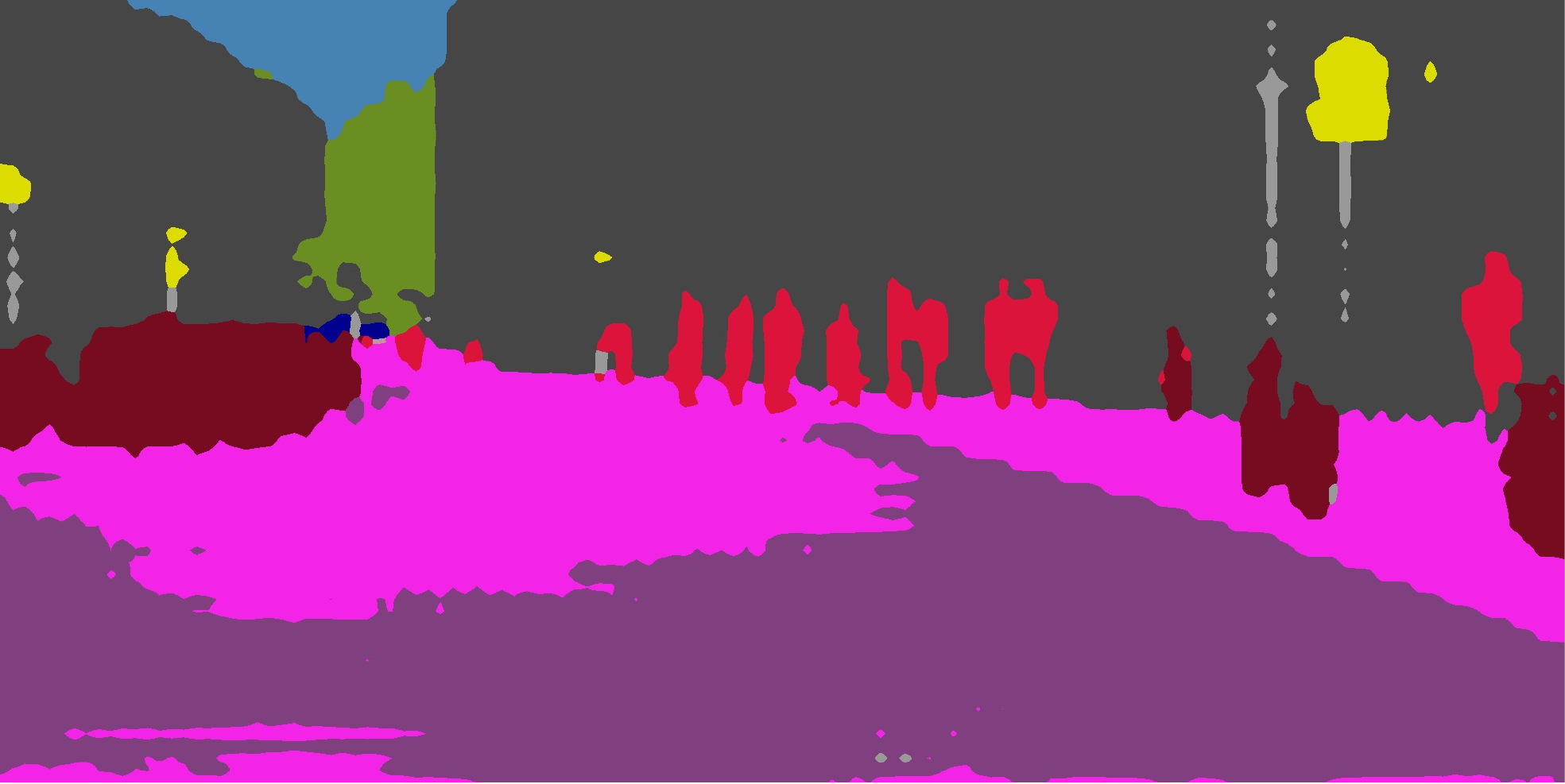} &
    \includegraphics[width=0.15\linewidth]{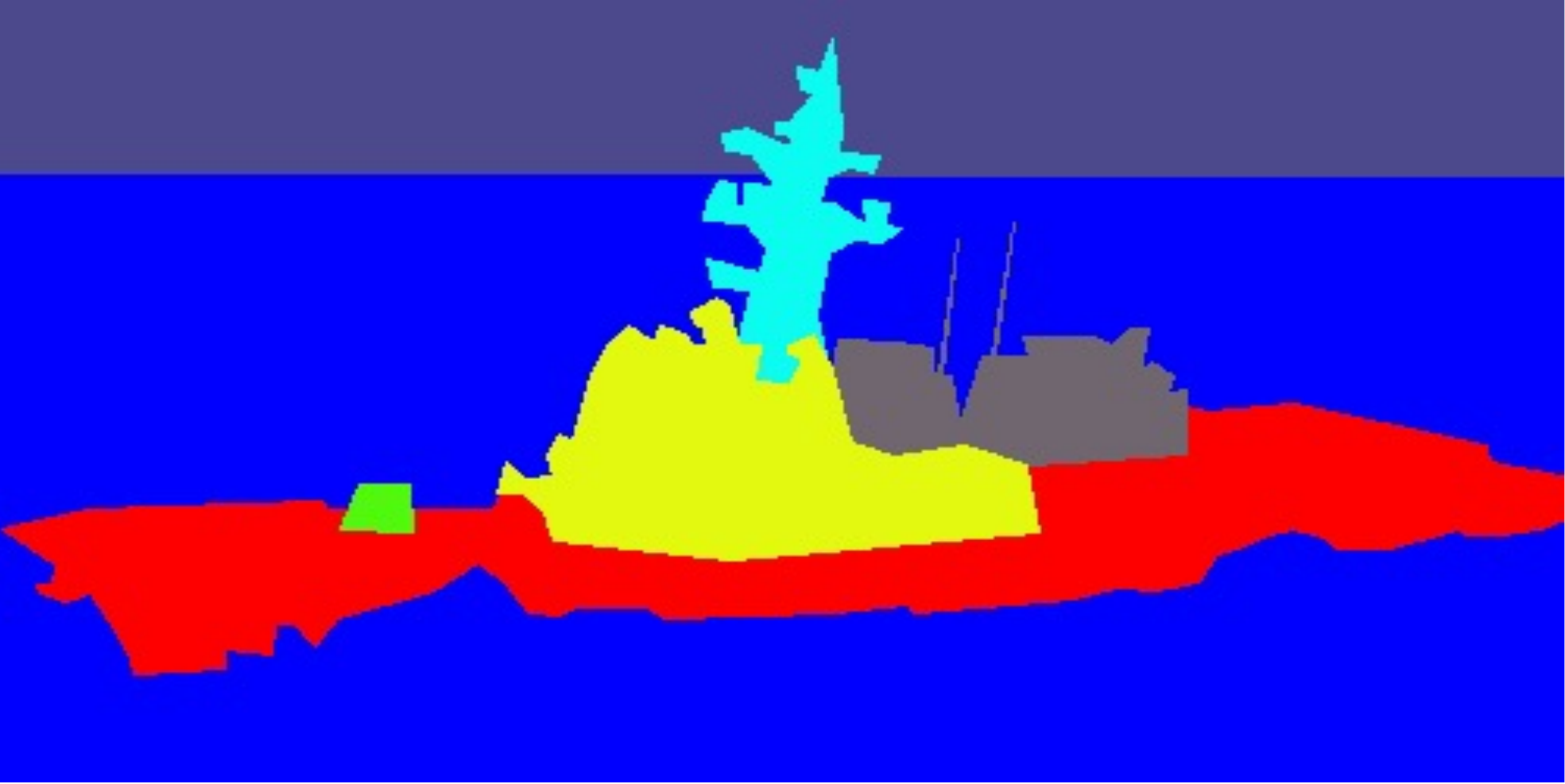} &
    \includegraphics[width=0.15\linewidth]{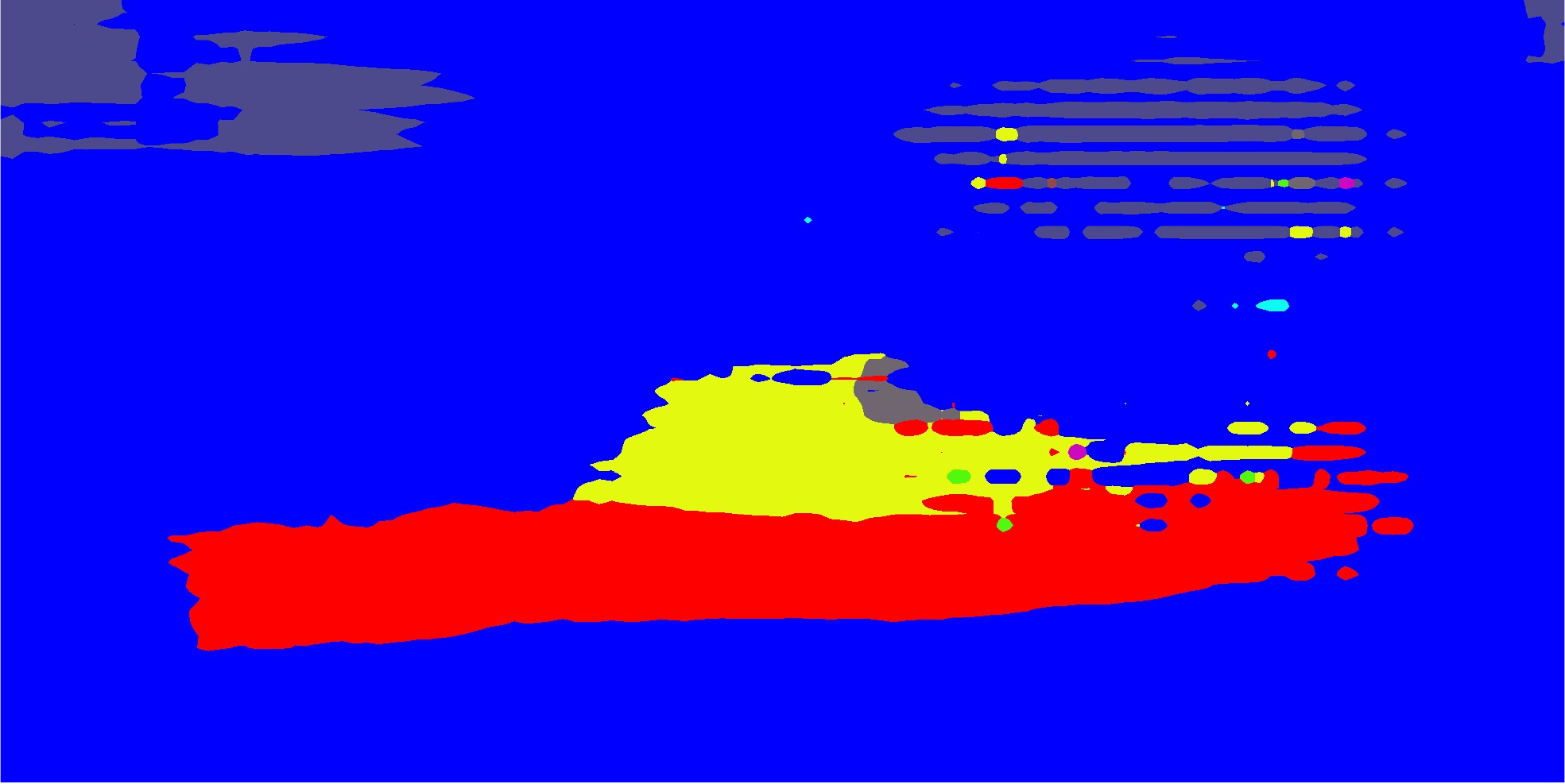} &
    \includegraphics[width=0.15\linewidth]{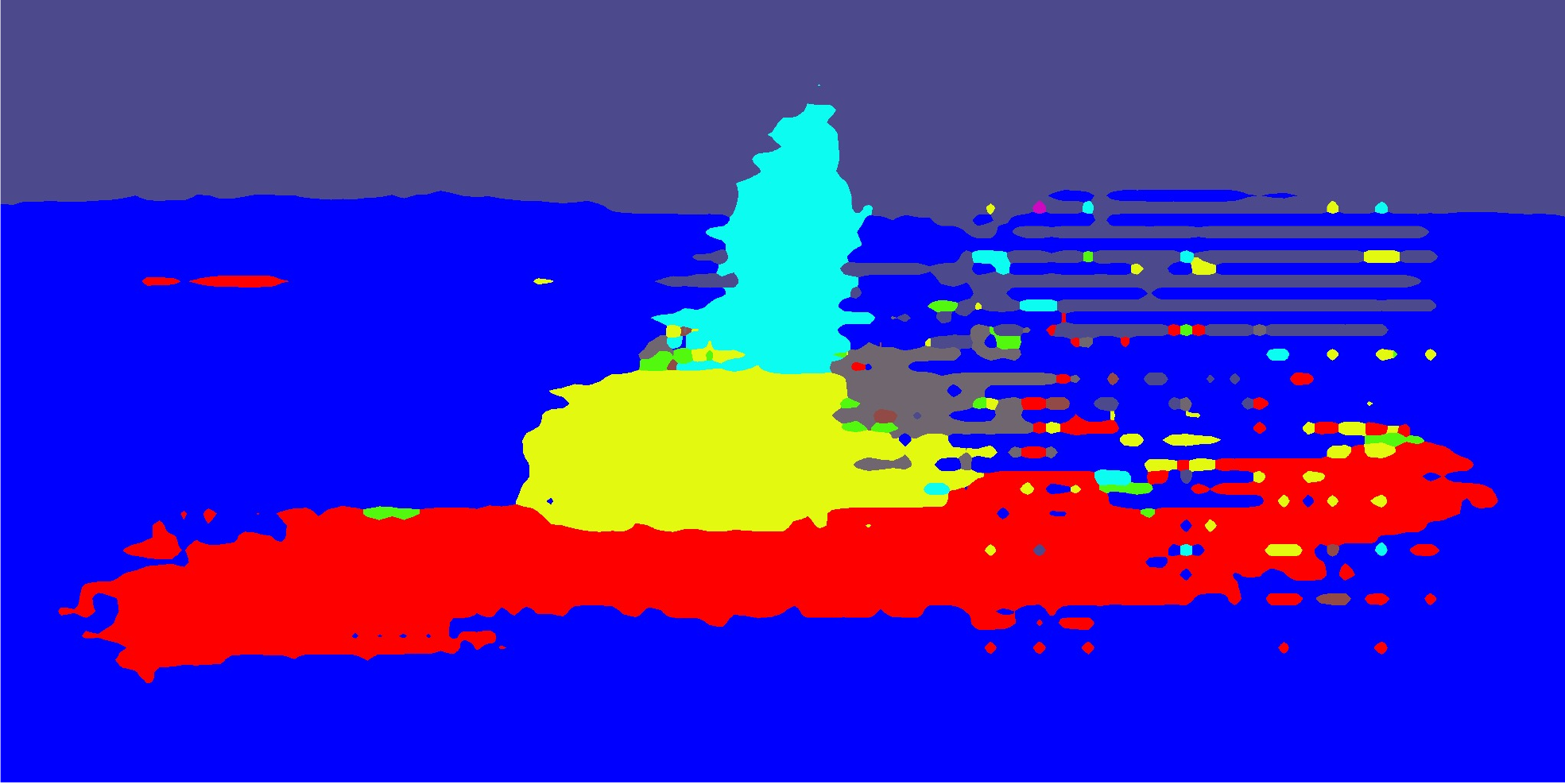} & \hspace{+0.5mm} \scriptsize\rotatebox{90}{\,Baseline} \\
    \vspace{+0.03mm}& \vspace{+0.03mm}& \vspace{+0.03mm}& \vspace{+0.03mm}& \vspace{+0.03mm}& \vspace{+0.03mm}& \\
    \vspace{-2mm}\tiny{Target image / GT} & \tiny{100} & \tiny{500} & \tiny{Target image / GT} & \tiny{100} & \tiny{500} & \\
    \end{tabular}}
    \caption{
    Semantic segmentation result comparisons between baseline and our methods on various datasets.\vspace{-2mm}}
    \label{fig:fig5} 
\end{figure*} 

\noindent\textbf{SYNTHIA to Cityscapes.} Similar to the GTA5 $\rightarrow$ Cityscapes scenario, our SDSS method shows consistent performance improvement in the SYNTHIA $\rightarrow$ Cityscapes scenario as shown in \tabref{table:table1}{(b)}.
These results demonstrate the generalization capability of the proposed SDSS method across various synthetic datasets.

\begin{figure}[ht]
    \caption{mIoU-Iteration analysis on \textit{Ocean Ship} (synthetic $\rightarrow$ real).
    }
    \centering
    \includegraphics[width=0.65\linewidth]{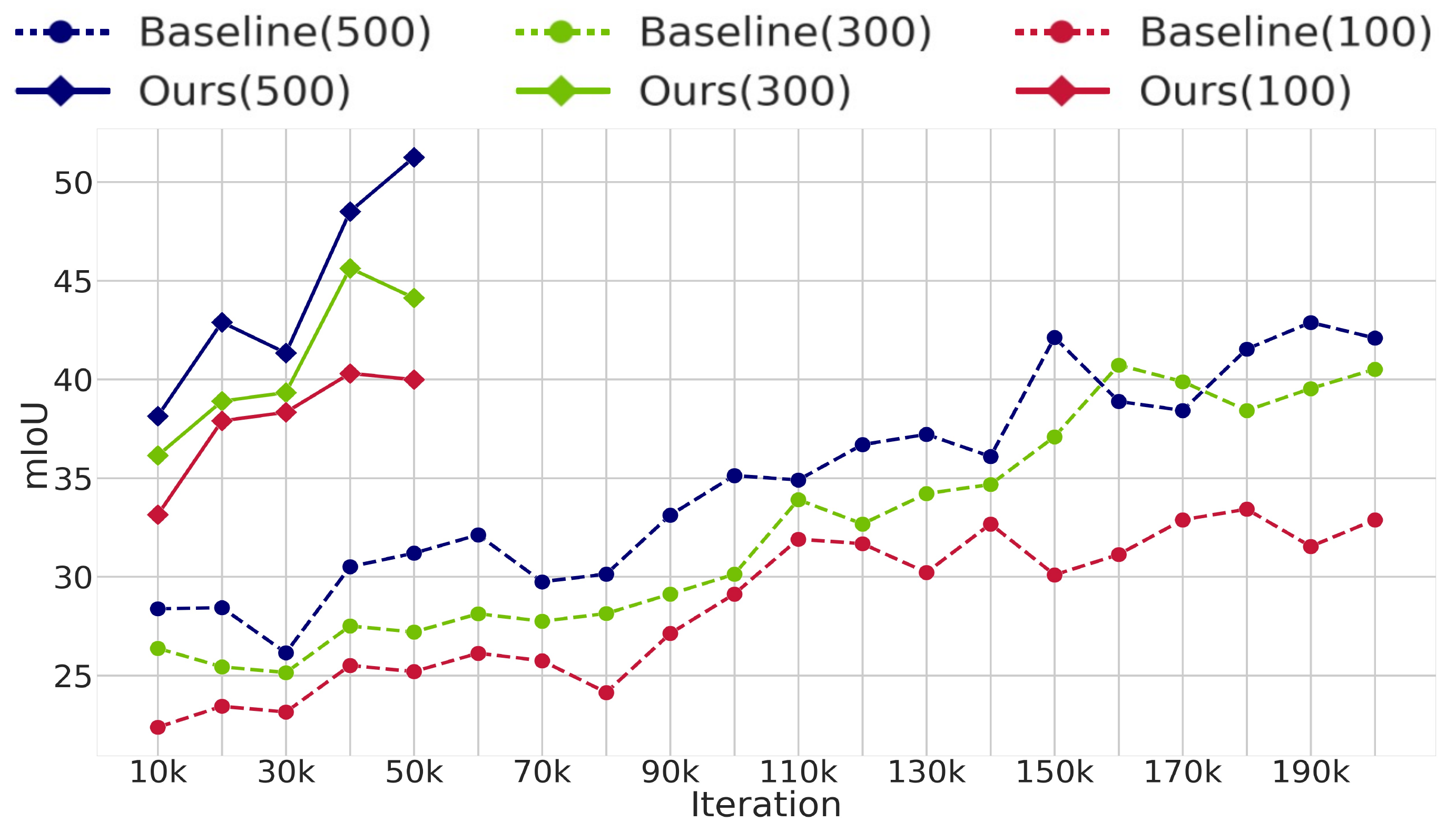}
    \label{fig:ocean_g}
\end{figure}

\noindent\textbf{\textit{Ocean Ship} (synthetic to real).} 
Our \textit{Ocean Ship} dataset provides large-scale source domain images, and the difference in the number of images between the source and target domains is large.
However, generating several source data and using the entire set for training is inefficient (i.e, the more training data, the longer the training time).
Therefore, we applied the SDSS method to our \textit{Ocean Ship }dataset to efficiently utilize synthetic data and to reduce the computational cost.
\tabref{table:ocean_table} shows the experimental results on our new \textit{Ocean Ship} dataset. 
Our method achieved a performance improvement of up to 9.13 mIoU, despite the reduced amount of training data compared to the baseline.
\figshortref{fig:ocean_g} shows the mIoU-Iteration analysis. Deep networks with SDSS achieves higher performance with fewer samples and less training time.
These results indicate that the SDSS method works better in real-world scenarios where synthetic data can be generated indefinitely.

\figshortref{fig:fig5} shows the qualitative segmentation performance comparisons between baseline and our methods with 100 and 500 available target GT labels.
Our method consistently outperforms the baseline without SDSS, although the number of available target domain data is the same.
In addition, performance is further improved with the increased number of available target data (\cf \figshortref{fig:fig5} (a) and  (b)).
Similar results are obtained with our \textit{Ocean Ship} dataset as shown in \figshortref{fig:fig5} (c).

\begin{table}[h]
 \centering
 \scriptsize
 \caption{Quantitative results on \textit{Ocean Ship} (synthetic $\rightarrow$ real).\vspace{-2mm}}
 \resizebox{.97\textwidth}{!}{
\begin{tabular}[b]{cc|ccc}
    \hline \hline
    \multicolumn{5}{c}{\textbf{\textit{Ocean Ship} (synthetic $\rightarrow$ real)}} \\ \hline
    \multirow{2}{*}{Type} & \multirow{2}{*}{Method} & \multicolumn{3}{c}{Number of labelled targets} \\
    & & 100 & 300 & 500 \\ \hline
    Supervised & Deeplab-v2  & 33.10 & 37.46 & 43.88 \\ \hline
    \multirow{2}{*}{SSDA} & Baseline & 35.67 & 40.72 & 42.12 \\
    & \cellcolor{Gray} Baseline + Ours  & \cellcolor{Gray}\textbf{40.30(+4.63)} & \cellcolor{Gray}\textbf{45.63(+4.91)} &                 \cellcolor{Gray}\textbf{51.25(+9.13)} \\
    \hline
    \multicolumn{2}{c|}{Number of labelled sources (Ours)} & 41,140 & 46,574 & 48,802 \\ \hline \hline
    \end{tabular}}
    \label{table:ocean_table}
\end{table}

\captionsetup[subfloat]{labelformat=empty}

\begin{table*}[!t]
 \centering
 \scriptsize
  \subfloat[\footnotesize (a) Pixel-level sampling]{%
    \resizebox{0.32\textwidth}{!}{%
      \def\arraystretch{1.05}
        \begin{tabular}[b]{cccc}
        \hline \hline
        \multirow{2}{*}{\# of Source Samples} & \multicolumn{3}{c}{Label Type} \\ \cline{2-4}
         & $\mathbf{y}^{s} \backslash \mathbf{y}^{c}$ & $\mathbf{y}^{s}$ & $\mathbf{y}^{c}$ \\
        \hline
        2,497 (10\%) & 7.60 & 33.71 & 39.97 \\ \hline
        7,490 (30\%) & 16.46 & 36.56 & 42.07 \\ \hline
        12,483 (50\%) & 20.28 & 38.47 & 42.14 \\ \hline
        17,476 (70\%) & 19.35 & 36.30 & \textbf{45.39} \\ \hline
        24,966 (100\%) & 15.37 & 36.92 & 44.56 \\ \hline \hline
        \end{tabular}
        \label{table:Pixel_eff}
    }
  } 
  \subfloat[\footnotesize (b) Image-level sampling]{%
    \resizebox{0.32\textwidth}{!}{%
      \def\arraystretch{1.17}
        \begin{tabular}[b]{ccc}
        \hline \hline
        Sampling Type   & \# of Source Samples & mIoU \\ \hline
        Top-10\% & {2,497} & 40.88    \\\hline
        Top-30\% & {7,490} & 42.48    \\\hline
        Top-50\% & 12,483 &43.71 \\ \hline
        Top-70\% & 17,476  & \textbf{45.42}    \\\hline
        100\% & 24,966 & 44.56 \\ \hline \hline
        \end{tabular}
        \label{table:Top_Bottom}
    }
  }  
  \subfloat[\footnotesize (c) Threshold of $\tau_{C}$]{%
    \resizebox{0.32\textwidth}{!}{%
    \def\arraystretch{1.17}
        \begin{tabular}[b]{ccc}
        \hline \hline
        Sampling Type & \# of Source Samples & mIoU \\ \hline
        $\tau_{C} > 0.1$ & 24,939 & 44.63    \\ \hline
        $\tau_{C} > 0.2$ & 24,018 & {45.11}    \\ \hline
        $\tau_{C} > 0.3$ & 19,695 & \textbf{45.48}    \\ \hline
        $\tau_{C} > 0.4$ & 7,275 & {42.46}    \\ \hline
        $\tau_{C} > 0.5$ & 936 & 33.93 \\ \hline \hline
        \end{tabular}
        \label{table:eff_thr}
    }
  }
\caption{Empirical experiment results on the effect of (a) pixel-level sampling, (b) image-level sampling, and (c) Threshold of $\tau_{C}$, which constitute our SDSS.}
\end{table*}

\begin{table*}[!t]
\resizebox{.98\textwidth}{!}{
    \centering
    \def\arraystretch{1.2}
    \begin{tabular}{c|cccccccccccccccccccc}
    \hline \hline
     \multicolumn{21}{c}{\textbf{GTA5} $\rightarrow$ \textbf{Cityscapes}}\\ \hline
    Type & \rotatebox[origin=c]{90}{road} & \rotatebox[origin=c]{90}{sidewalk} & \rotatebox[origin=c]{90}{building} & \rotatebox[origin=c]{90}{fence} & \rotatebox[origin=c]{90}{wall} & \rotatebox[origin=c]{90}{pole} & \rotatebox[origin=c]{90}{light} & \rotatebox[origin=c]{90}{sign} & \rotatebox[origin=c]{90}{veg.} & \rotatebox[origin=c]{90}{terrain} & \rotatebox[origin=c]{90}{sky} & \rotatebox[origin=c]{90}{person} & \rotatebox[origin=c]{90}{rider} & \rotatebox[origin=c]{90}{car} & \rotatebox[origin=c]{90}{truck} & \rotatebox[origin=c]{90}{bus} & \rotatebox[origin=c]{90}{train} & \rotatebox[origin=c]{90}{motocycle} & \rotatebox[origin=c]{90}{bicycle} & mIoU\\
    \hline
    w/o class balance & 71.75 & 40.62
    & 72.65 & 18.88
    &18.88&26.43&19.37&19.81&80.15&14.66&79.78&49.84&6.85&52.85&15.32&24.02&0.0&0.0&18.17 & 33.15\\
    \hline
    w/ class balance & \textbf{79.98} & \textbf{42.20} & \textbf{78.21} & \textbf{24.18} & \textbf{29.46} & \textbf{28.16} & \textbf{22.81} & \textbf{25.18} & \textbf{81.87} & \textbf{21.64} & \textbf{79.83} & \textbf{54.88} & \textbf{26.14} & \textbf{63.94} & \textbf{15.98} & \textbf{43.52} & \textbf{0.5} & \textbf{22.37} & \textbf{36.35} & \textbf{40.88} \\
    \hline\hline 

    \end{tabular}
    \caption{
    Quantitative comparison of image-level sampling with and without class balance.
    Top-10\% of source data is sampled based on $\tau_{C}$ with and without considering balanced class distribution in images for the training.}
    \label{table:class_balance}
}
\end{table*}

\begin{table}
\resizebox{.98\textwidth}{!}{
    \centering
    \def\arraystretch{1.2}
    \begin{tabular}{c|ccccccccc}
    \hline \hline
     \multicolumn{10}{c}{\textbf{Cityscapes} $\rightarrow$ \textbf{Foggy Cityscapes}}\\ \hline
    Type & person & rider & car & truck & bus & train & motocycle & bicycle & mAP\\
    \hline
    Faster R-CNN~\cite{ren2015faster} (w/o sampling) & 24.1 & 33.1 & 34.3 & 4.1 & 22.3 & 3.0 & 15.3 &26.5 & 20.3\\
    \hline
    Faster R-CNN~\cite{ren2015faster} (w/ sampling) & \textbf{25.0} & 31.0 & \textbf{36.3} & \textbf{20.1} & \textbf{29.3} & \textbf{18.2} & \textbf{19.0} & \textbf{27.1} & \textbf{25.8}\\   
    \hline\hline 
    \end{tabular}
    \caption{
    Experimental results applying our sampling method in Cityscapes to Foggy Cityscapes~\cite{sakaridis2018model} domain adaptation for object detection.
    (Instead of pixel-leval sampling, instance-level sampling is used.)
    Note that instance-level sampling is used instead of pixel-level sampling.
    }
    \label{table:detection}
}
\end{table}

\subsection{Ablation studies}
To further validate the proposed SDSS, we presented ablation studies on the effects of the pixel-level sampling, image-level sampling, and class balancing with detailed analyses.
The GTA5 $\rightarrow$ Cityscapes dataset was used for the experiments.

\noindent\textbf{Effect of pixel-level sampling.} To verify the effect of the pixel-level sampling, we compared segmentation performances on the target domain of the models trained with and without pixel-level sampling.
We randomly selected 500 images from the target domain training set to pre-train the network that will be used for the sampling on the source dataset.
Then, we randomly selected various source domain images to train the segmentation network with and without pixel-level sampling.
Note that only source domain sampled data were used for the training in this experiment.
Tabel \textcolor{red}{3} (a) shows the performance comparison results.
With exactly the same subsets of source domain images, models trained with our pixel-level sampling (i.e, $\mathbf{y}^{c}$) showed a significant performance improvement up to $9.09$ mIoU over models trained with raw GT labels without sampling (i.e, $\mathbf{y}^{s}$).
We also trained the model with labels excluded by the pixel-level sampling (i.e, $\mathbf{y}^{s} \backslash \mathbf{y}^{c}$), and it showed a substantial performance degradation.
These results indicate that our pixel-level sampling method correctly removes samples that are irrelevant for the target domain performance improvement as expected.

\noindent\textbf{Effect of image-level sampling.} We also verified the effect of the image-level sampling by training the models with top $k$ percent source domain data based on our scoring $\tau_{C}$.
Table \textcolor{red}{3} (b)  shows the performance comparisons with $k \in \{10, 30, 50, 70\}$.
%
Models trained on top-$k$ images consistently outperform models trained on images randomly sampled in Table \textcolor{red}{3} (a) for all $k$ values, as expected.
These results indicate that the proposed image-level sampling scheme can distinguish samples helpful for the training.
Table \textcolor{red}{3} (c) shows the performance comparisons with various threshold values for $\tau_{C}$.
The lower the $\tau_{C}$, the larger the number of images for the training.
However, segmentation performance is not proportionally increased.
If $\tau_{C}$ is too low, then several images with little information will be used for the training.
Contrarily, many useful images will be excluded from the training data if $\tau_{C}$ is too high.
Therefore, properly setting the threshold is important for improving performance in the target domain.
In our experiments, $\mathbf{\tau}_{C} > 0.3$ showed the best performance, so it was used in all the experiments.

\noindent\textbf{Effect of class balance on $\tau_{C}$.} The proposed class balance score $\tau_{C}$ considers the correctness ratio and pixel ratio of each class in an image (c.f, \secref{subsec:image_level}).
To analyse the effect of the class balance term in \eqnref{eqn:image_score} (i.e, $1 - n_{k}^{class} / n^{image}$), we conducted image-level sampling based on $\tau_{C}$ with and without the class balance term.
\tabfref{table:class_balance} shows the performance comparison of the models trained with and without the class balance term.
The model with the class balance term consistently outperformed its counterpart for all classes.
In particular, performance improvement is significant for the classes typically occupying small areas in an image such as motorcycles, bicycles, buses, riders, and terrain.
Therefore, our SDSS with class balance consideration is effective in improving segmentation performance for minor classes in datasets.

\section{Discussion \& future work}

In this study, we investigated a method to efficiently improve the performance of domain adaptation by sampling source data for semantic segmentation. 
Our motivation can be applied in domain adaptation of various tasks other than semantic segmentation.
However, pixel-level sampling is dependent on semantic segmentation.
Therefore, to apply our method to object detection, we adopt instance-level sampling, which applies bounding box-level sampling within a single image, instead of pixel-level sampling, which depends on semantic segmentation.
For example, if the target knowledge network correctly predicts $c$ out of $t$ GT bounding boxes and classes in an image, then the supervision of $t - c$ bounding boxes would not be used for the training.

The results presented in \tabref{table:detection} indicate that using source domain data sampling is also effective for domain adaptation in object detection. The network used in the experiment and all hyperparameters follow~\cite{renNIPS15fasterrcnn}.
However, because our experimental results involve simply replacing pixel-level sampling with instance-level sampling rather than applying or optimizing all our sampling methods, we intend to further investigate how to efficiently apply other tasks in future work.

\section{Conclusion}
\label{sec:Conclusion}
In this paper, we proposed source domain subset sampling (SDSS) method for semi-supervised domain adaptation.
Our method effectively subsamples source domain data with the proposed pixel-level and image-level sampling strategies based on the knowledge learned from a small amount of available target domain labelled data.
To verify the effectiveness of our method, we further have constructed a new dataset called \textit{Ocean Ship} containing 500 real and 200K synthetic images with GT labels.
Experimental results on two public datasets and our dataset demonstrate that our method has brought significant segmentation performance improvement with a reduced number of source domain data in a shorter training time.

{\small
\bibliographystyle{ieee_fullname}
\bibliography{egbib}

\begin{thebibliography}{10}\itemsep=-1pt

\bibitem{alhaija2018augmented}
Hassan~Abu Alhaija, Siva~Karthik Mustikovela, Lars Mescheder, Andreas Geiger,
  and Carsten Rother.
\newblock Augmented reality meets computer vision: Efficient data generation
  for urban driving scenes.
\newblock {\em {Int'l Journal of Computer Vision (IJCV)}}, 126(9):961--972,
  2018.

\bibitem{chen2017deeplab}
Liang-Chieh Chen, George Papandreou, Iasonas Kokkinos, Kevin Murphy, and Alan~L
  Yuille.
\newblock Deeplab: Semantic image segmentation with deep convolutional nets,
  atrous convolution, and fully connected crfs.
\newblock {\em IEEE transactions on pattern analysis and machine intelligence},
  40(4):834--848, 2017.

\bibitem{chen2018encoder}
Liang-Chieh Chen, Yukun Zhu, George Papandreou, Florian Schroff, and Hartwig
  Adam.
\newblock Encoder-decoder with atrous separable convolution for semantic image
  segmentation.
\newblock In {\em Proceedings of the European conference on computer vision
  (ECCV)}, pages 801--818, 2018.

\bibitem{choi2020cars}
Sungha Choi, Joanne~T Kim, and Jaegul Choo.
\newblock Cars can't fly up in the sky: Improving urban-scene segmentation via
  height-driven attention networks.
\newblock In {\em Proceedings of the IEEE/CVF conference on computer vision and
  pattern recognition}, pages 9373--9383, 2020.

\bibitem{cordts2016cityscapes}
Marius Cordts, Mohamed Omran, Sebastian Ramos, Timo Rehfeld, Markus Enzweiler,
  Rodrigo Benenson, Uwe Franke, Stefan Roth, and Bernt Schiele.
\newblock The cityscapes dataset for semantic urban scene understanding.
\newblock In {\em {Proc. of Computer Vision and Pattern Recognition (CVPR)}},
  pages 3213--3223, 2016.

\bibitem{deng2009imagenet}
Jia Deng, Wei Dong, Richard Socher, Li-Jia Li, Kai Li, and Li Fei-Fei.
\newblock Imagenet: A large-scale hierarchical image database.
\newblock In {\em 2009 IEEE conference on computer vision and pattern
  recognition}, pages 248--255. Ieee, 2009.

\bibitem{geiger2012we}
Andreas Geiger, Philip Lenz, and Raquel Urtasun.
\newblock Are we ready for autonomous driving? the kitti vision benchmark
  suite.
\newblock In {\em 2012 IEEE conference on computer vision and pattern
  recognition}, pages 3354--3361. IEEE, 2012.

\bibitem{he2016deep}
Kaiming He, Xiangyu Zhang, Shaoqing Ren, and Jian Sun.
\newblock Deep residual learning for image recognition.
\newblock In {\em Proceedings of the IEEE conference on computer vision and
  pattern recognition}, pages 770--778, 2016.

\bibitem{44873}
Geoffrey Hinton, Oriol Vinyals, and Jeffrey Dean.
\newblock Distilling the knowledge in a neural network.
\newblock In {\em NIPS Deep Learning and Representation Learning Workshop},
  2015.

\bibitem{huang2019apolloscape}
Xinyu Huang, Peng Wang, Xinjing Cheng, Dingfu Zhou, Qichuan Geng, and Ruigang
  Yang.
\newblock The apolloscape open dataset for autonomous driving and its
  application.
\newblock {\em IEEE transactions on pattern analysis and machine intelligence},
  42(10):2702--2719, 2019.

\bibitem{huang2019ccnet}
Zilong Huang, Xinggang Wang, Lichao Huang, Chang Huang, Yunchao Wei, and Wenyu
  Liu.
\newblock Ccnet: Criss-cross attention for semantic segmentation.
\newblock In {\em Proceedings of the IEEE/CVF International Conference on
  Computer Vision}, pages 603--612, 2019.

\bibitem{jiang2020bidirectional}
Pin Jiang, Aming Wu, Yahong Han, Yunfeng Shao, Meiyu Qi, and Bingshuai Li.
\newblock Bidirectional adversarial training for semi-supervised domain
  adaptation.
\newblock In {\em IJCAI}, pages 934--940, 2020.

\bibitem{kim2020attract}
Taekyung Kim and Changick Kim.
\newblock Attract, perturb, and explore: Learning a feature alignment network
  for semi-supervised domain adaptation.
\newblock In {\em European Conference on Computer Vision}, pages 591--607.
  Springer, 2020.

\bibitem{li2019bidirectional}
Yunsheng Li, Lu Yuan, and Nuno Vasconcelos.
\newblock Bidirectional learning for domain adaptation of semantic
  segmentation.
\newblock In {\em Proceedings of the IEEE/CVF Conference on Computer Vision and
  Pattern Recognition}, pages 6936--6945, 2019.

\bibitem{lin2014microsoft}
Tsung-Yi Lin, Michael Maire, Serge Belongie, James Hays, Pietro Perona, Deva
  Ramanan, Piotr Doll{\'a}r, and C~Lawrence Zitnick.
\newblock Microsoft coco: Common objects in context.
\newblock In {\em {Proc. of European Conf. on Computer Vision (ECCV)}}, pages
  740--755. Springer, 2014.

\bibitem{long2015fully}
Jonathan Long, Evan Shelhamer, and Trevor Darrell.
\newblock Fully convolutional networks for semantic segmentation.
\newblock In {\em Proceedings of the IEEE conference on computer vision and
  pattern recognition}, pages 3431--3440, 2015.

\bibitem{mayer2020adversarial}
Christoph Mayer and Radu Timofte.
\newblock Adversarial sampling for active learning.
\newblock In {\em Proceedings of the IEEE/CVF Winter Conference on Applications
  of Computer Vision}, pages 3071--3079, 2020.

\bibitem{mei2020instance}
Ke Mei, Chuang Zhu, Jiaqi Zou, and Shanghang Zhang.
\newblock Instance adaptive self-training for unsupervised domain adaptation.
\newblock {\em in European Conference on Computer Vision (ECCV)}, 2020.

\bibitem{minaee2021image}
Shervin Minaee, Yuri~Y Boykov, Fatih Porikli, Antonio~J Plaza, Nasser
  Kehtarnavaz, and Demetri Terzopoulos.
\newblock Image segmentation using deep learning: A survey.
\newblock {\em IEEE Transactions on Pattern Analysis and Machine Intelligence},
  2021.

\bibitem{moosbauer2019benchmark}
Sebastian Moosbauer, Daniel Konig, Jens Jakel, and Michael Teutsch.
\newblock A benchmark for deep learning based object detection in maritime
  environments.
\newblock In {\em Proceedings of the IEEE/CVF Conference on Computer Vision and
  Pattern Recognition Workshops}, pages 0--0, 2019.

\bibitem{musto2020semantically}
Luigi Musto and Andrea Zinelli.
\newblock Semantically adaptive image-to-image translation for domain
  adaptation of semantic segmentation.
\newblock {\em in BMVC2020}, 2020.

\bibitem{phuong2019distillation}
Mary Phuong and Christoph~H Lampert.
\newblock Distillation-based training for multi-exit architectures.
\newblock In {\em Proceedings of the IEEE/CVF International Conference on
  Computer Vision}, pages 1355--1364, 2019.

\bibitem{ren2015faster}
Shaoqing Ren, Kaiming He, Ross Girshick, and Jian Sun.
\newblock Faster r-cnn: Towards real-time object detection with region proposal
  networks.
\newblock {\em Advances in neural information processing systems}, 28:91--99,
  2015.

\bibitem{renNIPS15fasterrcnn}
Shaoqing Ren, Kaiming He, Ross Girshick, and Jian Sun.
\newblock Faster {R-CNN}: Towards real-time object detection with region
  proposal networks.
\newblock In {\em Advances in Neural Information Processing Systems ({NIPS})},
  2015.

\bibitem{richter2016playing}
Stephan~R Richter, Vibhav Vineet, Stefan Roth, and Vladlen Koltun.
\newblock Playing for data: Ground truth from computer games.
\newblock In {\em {Proc. of European Conf. on Computer Vision (ECCV)}}, pages
  102--118. Springer, 2016.

\bibitem{roitberg2021let}
Alina Roitberg, David Schneider, Aulia Djamal, Constantin Seibold, Simon
  Rei{\ss}, and Rainer Stiefelhagen.
\newblock Let's play for action: Recognizing activities of daily living by
  learning from life simulation video games.
\newblock {\em arXiv preprint arXiv:2107.05617}, 2021.

\bibitem{ronneberger2015u}
Olaf Ronneberger, Philipp Fischer, and Thomas Brox.
\newblock U-net: Convolutional networks for biomedical image segmentation.
\newblock In {\em International Conference on Medical image computing and
  computer-assisted intervention}, pages 234--241. Springer, 2015.

\bibitem{ros2016synthia}
German Ros, Laura Sellart, Joanna Materzynska, David Vazquez, and Antonio~M
  Lopez.
\newblock The synthia dataset: A large collection of synthetic images for
  semantic segmentation of urban scenes.
\newblock In {\em {Proc. of Computer Vision and Pattern Recognition (CVPR)}},
  pages 3234--3243, 2016.

\bibitem{saito2019semi}
Kuniaki Saito, Donghyun Kim, Stan Sclaroff, Trevor Darrell, and Kate Saenko.
\newblock Semi-supervised domain adaptation via minimax entropy.
\newblock In {\em Proceedings of the IEEE/CVF International Conference on
  Computer Vision}, pages 8050--8058, 2019.

\bibitem{sakaridis2018model}
Christos Sakaridis, Dengxin Dai, Simon Hecker, and Luc Van~Gool.
\newblock Model adaptation with synthetic and real data for semantic dense
  foggy scene understanding.
\newblock In {\em Proceedings of the European Conference on Computer Vision
  (ECCV)}, pages 687--704, 2018.

\bibitem{shao2018seaships}
Zhenfeng Shao, Wenjing Wu, Zhongyuan Wang, Wan Du, and Chengyuan Li.
\newblock Seaships: A large-scale precisely annotated dataset for ship
  detection.
\newblock {\em IEEE transactions on multimedia}, 20(10):2593--2604, 2018.

\bibitem{SunZJCXLMWLW19}
Ke Sun, Yang Zhao, Borui Jiang, Tianheng Cheng, Bin Xiao, Dong Liu, Yadong Mu,
  Xinggang Wang, Wenyu Liu, and Jingdong Wang.
\newblock High-resolution representations for labeling pixels and regions.
\newblock {\em CoRR}, abs/1904.04514, 2019.

\bibitem{touvron2021training}
Hugo Touvron, Matthieu Cord, Matthijs Douze, Francisco Massa, Alexandre
  Sablayrolles, and Herv{\'e} J{\'e}gou.
\newblock Training data-efficient image transformers \& distillation through
  attention.
\newblock In {\em International Conference on Machine Learning}, pages
  10347--10357. PMLR, 2021.

\bibitem{tsai2018learning}
Yi-Hsuan Tsai, Wei-Chih Hung, Samuel Schulter, Kihyuk Sohn, Ming-Hsuan Yang,
  and Manmohan Chandraker.
\newblock Learning to adapt structured output space for semantic segmentation.
\newblock In {\em Proceedings of the IEEE conference on computer vision and
  pattern recognition}, pages 7472--7481, 2018.

\bibitem{vu2019advent}
Tuan-Hung Vu, Himalaya Jain, Maxime Bucher, Matthieu Cord, and Patrick
  P{\'e}rez.
\newblock Advent: Adversarial entropy minimization for domain adaptation in
  semantic segmentation.
\newblock In {\em Proceedings of the IEEE/CVF Conference on Computer Vision and
  Pattern Recognition}, pages 2517--2526, 2019.

\bibitem{wang2020alleviating}
Zhonghao Wang, Yunchao Wei, Rogerio Feris, Jinjun Xiong, Wen-Mei Hwu, Thomas~S
  Huang, and Honghui Shi.
\newblock Alleviating semantic-level shift: A semi-supervised domain adaptation
  method for semantic segmentation.
\newblock In {\em Proceedings of the IEEE/CVF Conference on Computer Vision and
  Pattern Recognition Workshops}, pages 936--937, 2020.

\bibitem{wang2020differential}
Zhonghao Wang, Mo Yu, Yunchao Wei, Rogerio Feris, Jinjun Xiong, Wen-mei Hwu,
  Thomas~S Huang, and Honghui Shi.
\newblock Differential treatment for stuff and things: A simple unsupervised
  domain adaptation method for semantic segmentation.
\newblock In {\em Proceedings of the IEEE/CVF Conference on Computer Vision and
  Pattern Recognition}, pages 12635--12644, 2020.

\bibitem{wei2020circumventing}
Longhui Wei, An Xiao, Lingxi Xie, Xin Chen, Xiaopeng Zhang, and Qi Tian.
\newblock Circumventing outliers of autoaugment with knowledge distillation.
\newblock {\em arXiv preprint arXiv:2003.11342}, 2(8), 2020.

\bibitem{yang2021context}
Jinyu Yang, Weizhi An, Chaochao Yan, Peilin Zhao, and Junzhou Huang.
\newblock Context-aware domain adaptation in semantic segmentation.
\newblock In {\em Proceedings of the IEEE/CVF Winter Conference on Applications
  of Computer Vision}, pages 514--524, 2021.

\bibitem{yang2020fda}
Yanchao Yang and Stefano Soatto.
\newblock Fda: Fourier domain adaptation for semantic segmentation.
\newblock In {\em Proceedings of the IEEE/CVF Conference on Computer Vision and
  Pattern Recognition}, pages 4085--4095, 2020.

\bibitem{yoo2019learning}
Donggeun Yoo and In~So Kweon.
\newblock Learning loss for active learning.
\newblock In {\em Proceedings of the IEEE/CVF Conference on Computer Vision and
  Pattern Recognition}, pages 93--102, 2019.

\bibitem{yu2020bdd100k}
Fisher Yu, Haofeng Chen, Xin Wang, Wenqi Xian, Yingying Chen, Fangchen Liu,
  Vashisht Madhavan, and Trevor Darrell.
\newblock Bdd100k: A diverse driving dataset for heterogeneous multitask
  learning.
\newblock In {\em Proceedings of the IEEE/CVF conference on computer vision and
  pattern recognition}, pages 2636--2645, 2020.

\bibitem{zhang2020state}
Beichen Zhang, Liang Li, Shijie Yang, Shuhui Wang, Zheng-Jun Zha, and Qingming
  Huang.
\newblock State-relabeling adversarial active learning.
\newblock In {\em Proceedings of the IEEE/CVF Conference on Computer Vision and
  Pattern Recognition}, pages 8756--8765, 2020.

\bibitem{zhang2019category}
Qiming Zhang, Jing Zhang, Wei Liu, and Dacheng Tao.
\newblock Category anchor-guided unsupervised domain adaptation for semantic
  segmentation.
\newblock In {\em Advances in Neural Information Processing Systems}, pages
  433--443, 2019.

\bibitem{zhao2017pyramid}
Hengshuang Zhao, Jianping Shi, Xiaojuan Qi, Xiaogang Wang, and Jiaya Jia.
\newblock Pyramid scene parsing network.
\newblock In {\em Proceedings of the IEEE conference on computer vision and
  pattern recognition}, pages 2881--2890, 2017.

\bibitem{zou2018unsupervised}
Yang Zou, Zhiding Yu, BVK Kumar, and Jinsong Wang.
\newblock Unsupervised domain adaptation for semantic segmentation via
  class-balanced self-training.
\newblock In {\em Proceedings of the European conference on computer vision
  (ECCV)}, pages 289--305, 2018.

\end{thebibliography}
}

\end{document}